\documentclass{article}

\PassOptionsToPackage{numbers, compress}{natbib}

\usepackage[preprint]{preprint_style}

\usepackage[utf8]{inputenc} 
\usepackage[T1]{fontenc}    
\usepackage{hyperref}       
\usepackage{url}            
\usepackage{fontawesome5}   
\usepackage{booktabs}       
\usepackage{amsfonts}       
\usepackage{nicefrac}       
\usepackage{microtype}      
\usepackage[table]{xcolor}  
\usepackage{amsmath}
\usepackage{graphicx}
\usepackage{subcaption}
\usepackage{multirow}
\usepackage{wrapfig}
\usepackage[most]{tcolorbox}
\usepackage{listings}
\usepackage{float}
\usepackage{enumitem}
\tcbset{promptbox/.style={colback=blue!4,colframe=blue!35,breakable,boxrule=0.5pt,arc=2pt,left=5pt,right=5pt,top=4pt,bottom=4pt}}
\lstdefinestyle{pyappendix}{language=Python,basicstyle=\ttfamily\scriptsize,keywordstyle=\color{blue!60!black},commentstyle=\color{gray!70!black},stringstyle=\color{green!40!black},showstringspaces=false,breaklines=true,breakatwhitespace=true,columns=fullflexible,keepspaces=true,frame=single,framerule=0.3pt,rulecolor=\color{gray!40},backgroundcolor=\color{gray!4},xleftmargin=1.0em,xrightmargin=1.0em}

\newcommand{\benchName}{XL-DocBench}
\newcommand{\retainedQANum}{1,519}

\newcommand{\bestcell}[1]{\cellcolor{green!18}\textbf{#1}}
\newcommand{\secondcell}[1]{\cellcolor{blue!12}#1}

\title{XL-DocBench: Benchmarking Evidence-Grounded Extra-Long Document Understanding}

\author{%
  \quad \textbf{Hongchen Wei}$^{1,{\dagger},{\ddagger}}$
  \quad \textbf{Yuanzhe Wang}$^{2,{\dagger},{\ddagger}}$
  \quad \textbf{Bei Liu}$^{2,*}$ 
  \quad \textbf{Yifan Yang}$^{2}$ \\
  \quad \textbf{Qi Dai}$^{2}$
  \quad \textbf{Ruichun Ma}$^{2}$
  \quad \textbf{Kai Qiu}$^{2}$
  \quad \textbf{Yunsheng Li}$^{2}$
  \quad \textbf{Dongdong Chen}$^{2}$ \\
  \quad \textbf{Chong Luo}$^{2}$
  \quad \textbf{Zhenzhong Chen}$^{1}$
  \quad \textbf{Baining Guo}$^{2}$ \\[2ex]
  $^1$Wuhan University \quad $^2$Microsoft 
}

\begin{document}

\maketitle
\begingroup
\renewcommand\thefootnote{}
\footnotetext{$^\dagger$ Equal contribution. $^\ddagger$ Work done during an internship at MSRA. $^*$ Project leader.}
\endgroup

\begin{abstract}
Real-world document tasks often ask professionals to answer questions from annual reports, regulations, clinical guidelines, and technical manuals that span hundreds or thousands of pages. Some questions also require comparing related reports.
Reliable long-document understanding is therefore a prerequisite for using LLMs in compliance, clinical, financial, and engineering workflows, where decisions must be traceable to specific evidence pages and the cost of an unsupported answer is high -- yet most existing benchmarks still measure short-context or single-page QA.
We introduce \textbf{XL-DocBench}, a fully human-verified benchmark for extra-long document understanding, with \retainedQANum\ retained questions from six professional domains and contexts up to 2,303 pages.
XL-DocBench goes beyond page-level lookup. 1,103 examples (72.6\%) use multiple evidence pages. The final set also includes 556 questions (36.6\%) that use tables, charts, or figures, and 165 questions (10.9\%) that require evidence from multiple documents.
Each question has one of twelve reasoning labels, expert-annotated evidence pages, a typed verification rule, and an answer format, including 218 \texttt{None}-answer cases.
We build the benchmark with a tree-guided synthesis pipeline followed by artifact filters and full verification by 194 human experts.
By coupling extra-long professional contexts with page-level evidence and typed rules, \benchName{} fills a gap left by prior single-page, short multi-page, or text-only long-context benchmarks, and lets future work attribute system failures to retrieval, evidence use, or rule following rather than to a single leaderboard score.
The results show that current systems still struggle with long contexts, multi-page evidence, and structured reasoning over professional documents.
Project homepage: \href{https://officeintelligence.github.io/xl-docbench/}{\faGlobe}.

\end{abstract}

\section{Introduction}
\label{sec:introduction}

Extra-long professional documents are common in real document work. Annual reports, regulations, clinical guidelines, standards, and technical manuals often span hundreds or thousands of pages, and some questions require comparing related files.
In practice, users rarely ask for isolated snippets. They ask whether a condition is satisfied, which entities should be included in a set, how values change across reports, or whether the documents provide enough support to answer at all.
These questions are common in professional workflows because decisions often depend on policies, exceptions, definitions, footnotes, and repeated measurements that are scattered across long files.
Answering such questions is not a single-page lookup. A system must find the right parts of the document, read text together with tables and charts, combine evidence from different places, and decide when the documents do not support an answer \cite{ding2025survey}.
\begin{table}[t]
\centering
\small
\setlength{\tabcolsep}{5pt}
\renewcommand{\arraystretch}{1.08}
\begin{minipage}{0.96\textwidth}
\centering
\caption{Comparison between our benchmark and previous DU datasets.
\textbf{Unans.}: unanswerable question.
\textbf{TXT/L/C/TAB/I}: pure text/generalized layout/chart/table/image.
\textbf{Cross-doc.}: whether answering requires aggregating evidence across multiple documents.
\textbf{Avg. Evidence Pages}: average number of evidence pages needed to answer the question.
\textbf{N/A (text-only)}: text-only benchmark without native page concept.}
\label{tab:benchmark-comparison}
\resizebox{\linewidth}{!}{%
\begin{tabular}{@{}ccccccccc@{}}
\toprule
\multirow{2}{*}{\textbf{Benchmarks}} &
\multirow{2}{*}{\textbf{Release}} &
\multicolumn{2}{c}{\textbf{Document}} &
\multicolumn{3}{c}{\textbf{Question type}} &
\multicolumn{2}{c}{\textbf{Answer Evidence}} \\
& & \textbf{\# Pages} & \textbf{\# Tokens} & \textbf{Cross-page (\%)} & \textbf{Cross-doc. (\%)} & \textbf{Unans. (\%)} & \textbf{Source} & \textbf{Avg. Evidence Pages} \\
\midrule
DocVQA \cite{mathew2021docvqa} & 2020-07 & 1.0 & 151.5 & $\times$ & $\times$ & $\times$ & TXT/L/C/TAB/I & 1.0 \\
ChartQA \cite{masry2022chartqa} & 2022-03 & 1.0 & 236.9 & $\times$ & $\times$ & $\times$ & C & 1.0 \\
InfoVQA \cite{mathew2022infographicvqa} & 2021-04 & 1.2 & 288.0 & $\times$ & $\times$ & $\times$ & L/C/TAB/I & 1.0 \\
TAT-QA \cite{zhu2021tat} & 2022-07 & 1.1 & 577.0 & $\times$ & $\times$ & $\times$ & TXT/TAB & 1.0 \\
VisualWebBench \cite{liu2024visualwebbench} & 2024-04 & 1.0 & 452.4 & $\times$ & $\times$ & $\times$ & L/I & 1.0 \\
MP-DocVQA \cite{tito2023hierarchical} & 2022-12 & 8.3 & 2026.6 & $\times$ & $\times$ & $\times$ & TXT/L/C/TAB/I & 1.0 \\
DUDE \cite{van2023document} & 2023-05 & 5.7 & 1831.5 & $\checkmark$ ($\ast$) & $\times$ & $\checkmark$ ($\ast$) & TXT/L/C/TAB/I & -- \\
SlideVQA \cite{tanaka2023slidevqa} & 2023-01 & 20.0 & 2030.5 & $\checkmark$ ($\ast$) & $\times$ & $\times$ & TXT/L/C/TAB/I & -- \\
MMLongBench-Doc \cite{ma2024mmlongbench} & 2024-07 & 47.5 & 21214.1 & $\checkmark$ (33.0\%) & $\times$ & $\checkmark$ & TXT/L/C/TAB/I & 1.88 \\
LongDocURL \cite{deng2025longdocurl} & 2025-07 & 85.6 & 43622.6 & $\checkmark$ (52.9\%) & $\times$ & $\checkmark$ & TXT/L/TAB/I & 1.53 \\
\midrule
XL-DocBench & 2026-04 & 297.1 & 227463.2 & $\checkmark$ (72.6\%) & $\checkmark$ & $\checkmark$  & TXT/TAB/C/I & 2.30 \\
\bottomrule
\end{tabular}
}
\end{minipage}
\end{table}

As shown in Table~\ref{tab:benchmark-comparison}, existing document-understanding benchmarks cover parts of this setting: layout parsing, single-page QA, short multi-page reasoning, and text-only long-context tasks.
These datasets have helped the field, but they often do not require a model to justify an answer with exact evidence pages across very long professional files.
They also leave open how systems behave when the correct response is to abstain because the needed support is missing.
A model can therefore score well on average while still failing at the step users need most: finding the right pages and using them under an explicit rule.
A useful benchmark should use realistic document lengths and labels tied to evidence pages. It should also include multimodal cases, cross-document cases, and cases where support is missing, while filtering out questions answerable without the documents.

We introduce XL-DocBench, a fully human-verified benchmark for extra-long document understanding. It contains \retainedQANum\ retained questions over long professional documents from six domains. The median document context is 211 pages, and the longest context spans 2,303 pages.
XL-DocBench includes both local and multi-page evidence. 1,103 examples (72.6\%) use multiple evidence pages. Another 556 questions (36.6\%) involve tables, charts, or figures, 165 questions (10.9\%) are cross-document, and 218 examples use \texttt{None} as the answer format.
Questions are grouped into twelve reasoning types across three tiers, so we can see which reasoning steps fail instead of relying only on aggregate accuracy.
These labels help separate errors in finding evidence, tracking sets, comparing values, aggregating information, and recognizing missing support.

XL-DocBench is built with a tree-guided synthesis pipeline followed by artifact filters and full verification by 194 human experts. The automatic checks operate at the chapter or section level; experts then locate the exact supporting pages and quotes, verify the answer, and repair or remove ambiguous items.
This design separates generation at scale from final evidence checking: the pipeline proposes difficult candidates, but the released labels come from human review over the full documents.
Each retained example includes expert-annotated evidence pages, a typed verification rule, an answer format, and checks for support, completeness, and ambiguity.
Evaluations of proprietary VLMs, open-weight VLMs, and retrieval-augmented agent systems show that current systems still make many errors on this task.

Our main contributions are:
\begin{itemize}
    \item We present XL-DocBench, a fully human-verified benchmark for extra-long document understanding: \retainedQANum\ retained questions, six professional domains, twelve reasoning types, and diagnostic multimodal, \texttt{None}-answer, and cross-document slices.
    \item We introduce a tree-guided construction protocol with chapter/section-level path-dependency checks, typed verification rules, artifact filters, and page-level evidence annotation by 194 human experts.
    \item We provide evaluation results and breakdowns by document length, evidence span, reasoning type, document type, modality, and cross-document scope.
\end{itemize}

\section{Related Work}
\label{sec:related_work}

\textbf{Document Understanding Models and Methods.}
Document understanding methods have moved from layout-aware text models to multimodal readers and retrieval-augmented agents.
LayoutLM-style models combine OCR with two-dimensional layout embeddings~\cite{xu2020layoutlm,xu2021layoutlmv2,huang2022layoutlmv3,wang2022lilt,appalaraju2021docformer}, while end-to-end document readers such as Donut~\cite{kim2021donut} remove explicit OCR by decoding from page images.
Document-specialized and general-purpose VLMs now support higher resolution and longer context, including mPLUG-DocOwl~\cite{ye2023mplug,hu2025mplug}, TextMonkey~\cite{liu2026textmonkey}, Gemini~\cite{Gemini3}, Qwen3.5~\cite{Qwen35}, InternVL3.5~\cite{wang2025internvl3}, and Pixtral~\cite{agrawal2024pixtral}.
For hundreds-page documents, however, single-pass reading often fails or exceeds the model context window. Recent methods use visual page retrieval~\cite{faysse2024colpali,ma2025towards,yu2024visrag}, document RAG systems~\cite{cho2024m3docrag,tanaka2025vdocrag}, and agents that iteratively search, read, and accumulate evidence~\cite{han2025mdocagent,jain2025simpledoc,li2026deepread}.
These systems need evaluation on navigation and evidence use across long documents, not just on page reading.

\textbf{Document Understanding Benchmarks.}
Benchmarks have also moved from single-page visual QA to multi-page and long-context tasks.
DocVQA~\cite{mathew2021docvqa}, ChartQA~\cite{masry2022chartqa}, InfographicsVQA~\cite{mathew2022infographicvqa}, and TAT-QA~\cite{zhu2021tat} stress page-level visual, chart, infographic, and table-text reasoning; MP-DocVQA~\cite{tito2023hierarchical}, DUDE~\cite{van2023document}, and SlideVQA~\cite{tanaka2023slidevqa} extend to short multi-page inputs.
Recent long-document benchmarks such as MMLongBench-Doc~\cite{ma2024mmlongbench}, LongDocURL~\cite{deng2025longdocurl}, DocBench~\cite{zou2025docbench}, FinRAGBench-V~\cite{zhao2025finragbench}, M-LongDoc~\cite{ChiaCCSLAPB25}, MMDocBench~\cite{ZhuLNWWFWLC26}, and DocGenome~\cite{xia2024docgenome} increase length, domain specialization, or document formats. Text-only long-context benchmarks~\cite{shaham2022scrolls,bai2024longbench} test long-range reasoning without page-level visual structure.
What is still missing is a benchmark with realistic extra-long professional documents, evidence spread across pages or related documents, fine-grained reasoning labels, and construction checks that remove no-context or weakly supported questions.
XL-DocBench provides this combination with contexts up to 2{,}303 pages, 1,103 multi-page human-evidenced examples (72.6\% of the final set), a cross-document subset, twelve reasoning types, and full verification by 194 human experts.

\section{XL-DocBench}

This section describes how we build \benchName{} from long professional documents. The construction pipeline creates questions that need evidence from more than one local snippet while keeping each answer verifiable. We then summarize the dataset composition, reasoning labels, and validation protocol.

\subsection{Tree-Guided Coarse-to-Fine Construction}
\label{sec:pipeline}

Building \benchName{} cannot be done by simply asking annotators or models to write questions over long PDFs.
Manual annotation alone is hard to scale because experts must inspect hundreds or thousands of pages, understand domain terminology, find the relevant evidence, and check that the answer is well supported. Fully automatic generation has the opposite problem: it often produces local lookup questions, vague document-level questions, or answers without clear evidence.
We use \emph{human-guided, model-assisted construction}. Models propose candidates from selected document branches, automatic filters remove common errors, and human experts verify every retained example.
The pipeline is a data construction protocol, not a model architecture. It makes question generation easier to inspect and scale.

\begin{figure}[t]
    \centering
    \includegraphics[width=0.95\linewidth]{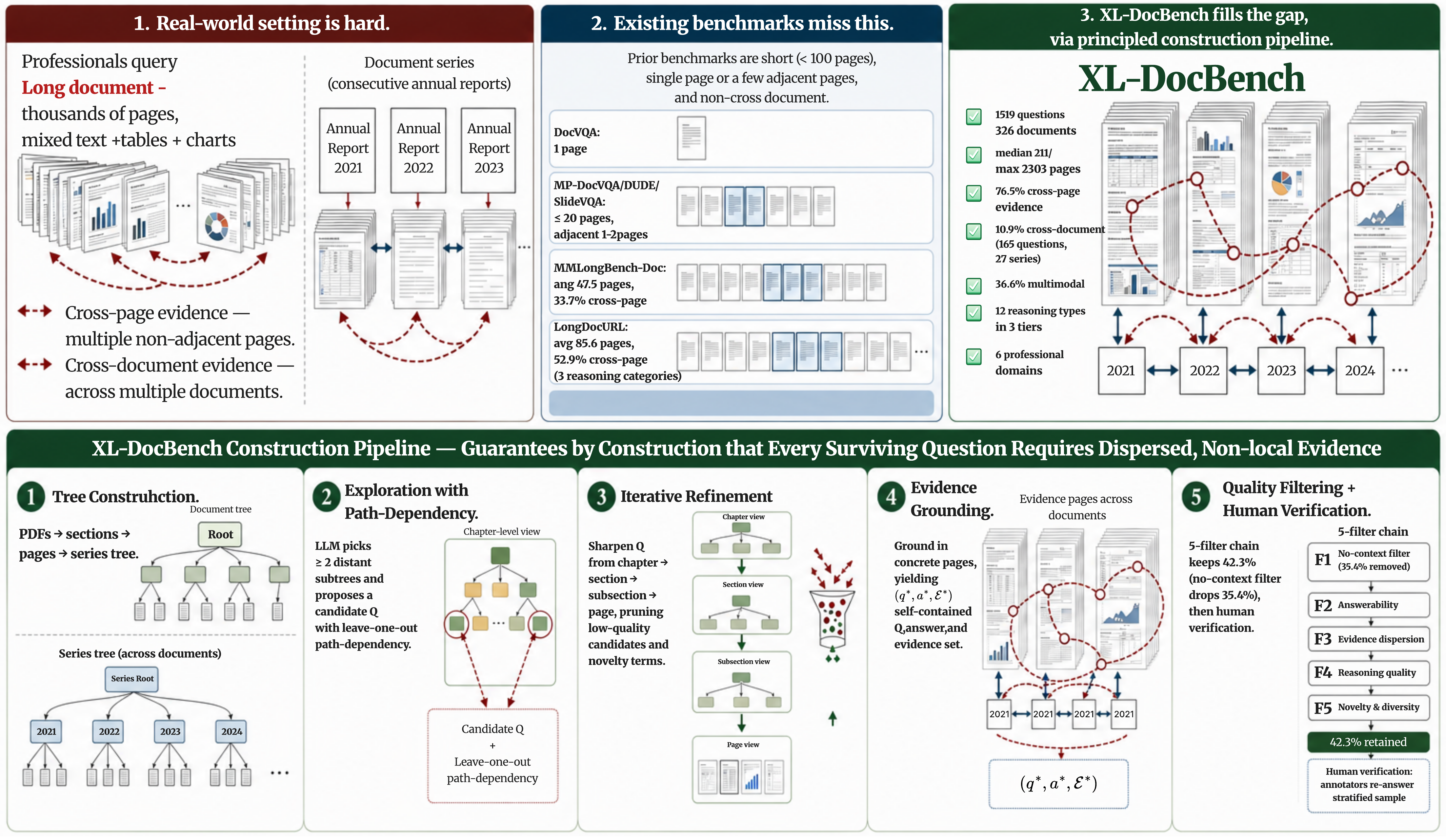}
    \caption{\textbf{XL-DocBench construction pipeline.} Long documents and document series are parsed into hierarchical trees. The pipeline generates candidate questions from multiple chapter/section branches, refines them with concrete content, adds candidate answers and verification rules, and filters artifacts. Full verification by 194 human experts identifies the final supporting pages and quotes, checks answer support and ambiguity, and removes or repairs invalid items.}
    \label{fig:pipeline}
\end{figure}

The pipeline starts with document structure and ends with human-verified examples. Each PDF is parsed into page images, Markdown, tables, and figure metadata, then organized into a document tree. We summarize tree nodes and sample chapter or section branches under domain and reasoning-type quotas. We then generate candidates from the branch summaries, test branch-level path-dependency, refine the questions with concrete content, add provisional answers and verification rules, filter artifacts, and ask human experts to annotate the exact supporting pages and quotes.

\textbf{Tree Construction.}
Inspired by PageIndex\footnote{\url{https://github.com/VectifyAI/PageIndex}}, we use a tree so the pipeline can move from broad sections to specific pages without prompting over the whole document at once.
Each PDF is represented as a hierarchy: internal nodes contain section summaries, and leaves correspond to pages. When headings are unreliable, we group pages using formatting cues.
For document series, we attach per-document trees to a shared root so cross-document candidates can be generated through the same interface.

\textbf{Exploration with Path-Dependency.}
A tree alone does not guarantee that a question needs evidence from multiple places. A question can look broad but still be answerable from one local excerpt.
We start at the coarsest level $\ell{=}0$ and check path-dependency over chapter- or section-level branches.
Given selected branches $\mathcal{U}$, the model proposes a candidate $q$ that should need information from each branch.
We use a leave-one-branch-out criterion:
\begin{equation}
\forall u\in\mathcal{U},\quad q \text{ is not answerable from } \mathcal{C}(\mathcal{U}\setminus\{u\}).
\end{equation}
This test does not prove that the final evidence set is minimal, and it does not identify the final evidence pages. It only filters out candidates that can be answered from one selected chapter or section.
An LLM judge applies the check to each leave-one-branch-out projection, and we keep candidates only when the missing branch removes necessary evidence. Human experts later annotate the final page-level evidence.
We also use quotas over twelve reasoning types so rare types such as counterfactual reasoning and reference chains are not lost during generation.

\textbf{Iterative Refinement.}
Candidates generated from summaries often involve the right branches, but they are too abstract to verify directly.
We expand selected branches one level at a time and rewrite the candidate with newly visible content:
\begin{equation}
q^{(\ell+1)} = \mathcal{R}_\ell\!\left(q^{(\ell)}, \{\mathrm{ch}(u): u\in\mathcal{U}^{(\ell)}\}\right).
\label{eq:iterative-refinement}
\end{equation}
where $\mathcal{U}^{(\ell)}$ is the current support frontier and $\mathrm{ch}(u)$ denotes its children.
After each step, a judge scores answerability, ambiguity, and reasoning depth. We remove low-quality candidates, keep strong candidates early, and use a novelty penalty to avoid repeating the same support combinations.

\textbf{Evidence Grounding.}
Refinement turns broad ideas into concrete questions, but the answer still needs to be checkable.
The automated stage proposes a self-contained question $q^\star$, a candidate answer $a^\star$, candidate support, and a typed verification rule; these remain provisional until expert verification.
Reasoning-type prompts handle cases such as counterfactual recalculation. A rewriting step then removes internal tree references.
Human experts then inspect the full source documents to mark the concrete evidence pages and supporting quotes that make the answer verifiable; this page-level annotation is the released ground truth.

\textbf{Quality Filtering.}
Even concrete examples can be invalid. Some are answerable from world knowledge, rely on unsupported or ambiguous evidence, use meaningless arithmetic, or contain internal tree references.
We filter these cases before and after expert verification. The final released subset retains 1,519 of the 3,550 synthetic candidates, and the no-context filter is the largest single rejection source.
The no-context filter presents each candidate question to a judge without any document content; candidates are rejected if the answer can be inferred from world knowledge, metadata, or wording leakage alone.
Filtering is not the final quality guarantee: every retained sample is verified by human experts for answer support, page-level evidence completeness, and ambiguity before release.
Additional filter details are provided in Appendix~\ref{app:pipeline_details}.

\subsection{Dataset Composition and Analysis}
\label{sec:dataset}

The pipeline in \S\ref{sec:pipeline} produces \benchName{} through several filtering stages. It starts with 3,550 synthetic candidates. Of these, 2,104 pass expert verification, and \retainedQANum\ remain in the final fully human-verified benchmark after additional difficulty-oriented filtering.
Table~\ref{tab:dataset-statistics} summarizes the final retained set.
These statistics define the main test conditions: long professional contexts, expert-annotated evidence, fine-grained reasoning labels, and subsets for multimodal, \texttt{None}-answer, long-span, and cross-document questions.

\begin{table}[t]
\centering
\scriptsize
\setlength{\tabcolsep}{3pt}
\renewcommand{\arraystretch}{0.93}
\caption{Dataset statistics. Funnel rows report annotation and filtering counts; other rows describe the final evaluation subset. Domain order: legal/finance/technical/medical/scientific/narrative; answer-format order: string/integer/float/none. Evidence-page rows distinguish old synthetic candidate pages from human-annotated evidence pages.}
\label{tab:dataset-statistics}
\resizebox{0.92\linewidth}{!}{%
\begin{tabular}{@{}lrlr@{}}
\toprule
\textbf{Statistic} & \textbf{Value} & \textbf{Statistic} & \textbf{Value} \\
\midrule
\multicolumn{4}{@{}l}{\emph{Construction funnel}} \\
All synthetic data & 3,550 & Human-verified candidates & 2,104 (59.3\%) \\
Final filtered subset & 1,519 & Candidate-to-final retention & 72.2\% \\
\addlinespace[1pt]
\multicolumn{4}{@{}l}{\emph{Final subset composition}} \\
Documents & 331 & Pages (total/avg/med/max) & 98,342/297.1/211/2,062 \\
Mean document tokens & 227,463.2 & Domains / reasoning types & 6/12 \\
Domain counts & 311/333/274/294/218/89 & Cross-doc Qs & 165 (10.9\%) \\
Multimodal evidence & 556 (36.6\%) & \texttt{None} answer & 218 (14.4\%) \\
Answer formats & 945/238/118/218 & Human experts & 194 \\
\addlinespace[1pt]
\multicolumn{4}{@{}l}{\emph{Final evidence-page statistics}} \\
Old synth. evidence. pages (avg/max) & 9.13/45 & Human-annotated evidence. pages (avg/max) & 2.30/24 \\
Multi-page human evidence. & 1,103 (76.4\%) & Human evidence. span (med/max) & 7/1,002 \\
\bottomrule
\end{tabular}%
}
\end{table}

\textbf{Corpus coverage.}
The retained benchmark covers six professional domains and long document contexts from authoritative public sources.
The median context is 211 pages and the mean is 297.1 pages; 51 examples exceed 1{,}000 pages.
Cross-document or series examples can span up to 2{,}303 pages, so they directly test retrieval across related documents.
Table~\ref{tab:dataset-statistics} gives the source and domain counts.

\begin{figure}[t]
    \centering
    \includegraphics[width=\linewidth]{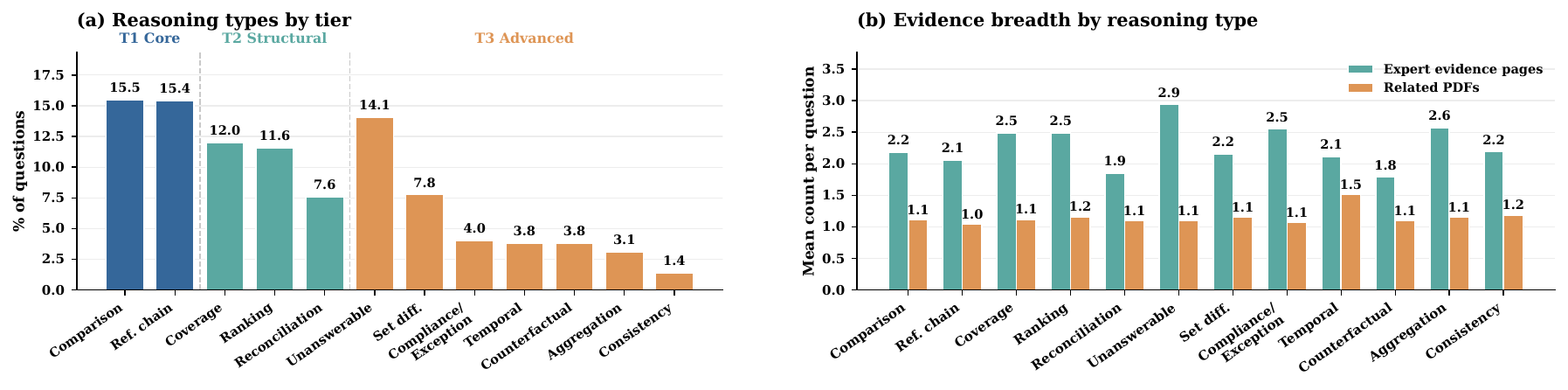}
    \caption{\textbf{Reasoning labels and evidence structure in the human-verified set.}
    \textbf{(a)} Distribution of the twelve reasoning types, colored by tier.
    \textbf{(b)} Mean number of expert-annotated evidence pages and related PDFs per reasoning type.}
    \label{fig:reasoning_panel}
\end{figure}

\textbf{Diagnostic reasoning taxonomy.}
For experiments, aggregate accuracy is not enough; we also need to know which reasoning steps fail.
\benchName{} assigns each question one of twelve mutually exclusive reasoning types, arranged into three tiers (Figure~\ref{fig:reasoning_panel}).
The taxonomy covers comparison, reference chain, ranking, coverage, reconciliation, set difference, unanswerable, temporal, compliance (including rule-special-case/exception checks), counterfactual, aggregation, and consistency.
The quotas in \S\ref{sec:pipeline} keep rare but important types instead of merging them into an ``other'' bucket.
Worked examples are given in Appendix~\ref{app:reasoning_case_studies}.

\begin{figure}[t]
    \centering
    \includegraphics[width=\linewidth]{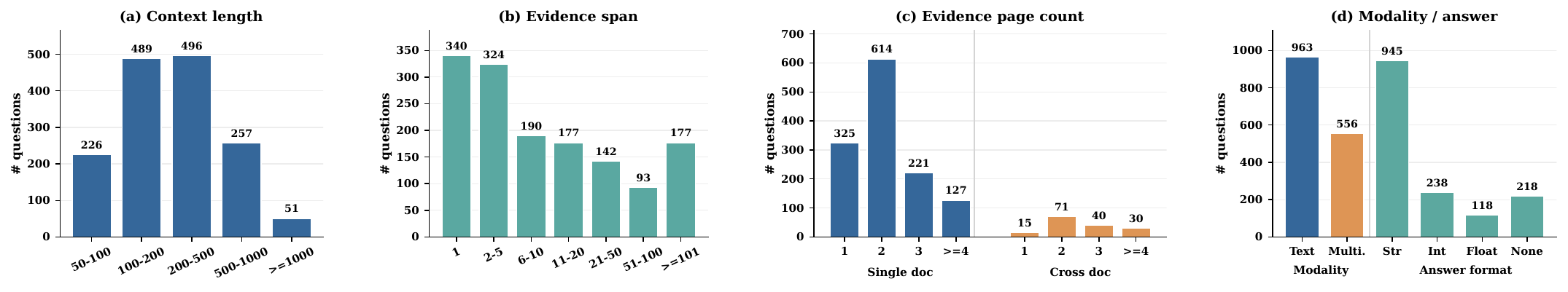}
    \caption{Composition of \benchName{} in the final human-verified set:
    (a) context length per example, summing PDFs for series questions;
    (b) evidence span from expert-annotated evidence pages;
    (c) evidence page count, separating single-page, within-document multi-page, and strict multi-PDF cases;
    (d) evidence modality and answer format.}
    \label{fig:stats}
\end{figure}

\textbf{Evidence grounding and span.}
The final human annotations show that the benchmark mixes local cases with many questions that need evidence from distant pages.
1,103 examples (72.6\%) require multiple evidence pages, while the remaining examples use a single annotated evidence page.
The median evidence span is 7 pages and the P95 span is 255 pages, so evidence localization over long contexts is a major part of the task.
Figure~\ref{fig:stats} summarizes context length, evidence span, evidence modality, and answer format.

\textbf{Diagnostic evaluation slices.}
Beyond aggregate statistics, \benchName{} includes three subsets that match common professional document use: evidence modality, document scope, and answer support.
The multimodal subset has 556 questions (36.6\%) that require tables, charts, or figures, so systems cannot rely only on plain OCR text.
The cross-document subset has 165 questions (10.9\%) that require evidence from related PDFs. We note that the cross-document label is defined by the \emph{set of documents the question is grounded in}, not by the count of pages on which the final supporting evidence happens to fall: a question may depend on multiple PDFs through context, definitions, or cross-references, yet its expert-annotated evidence pages may concentrate in only a few documents. The page-level evidence statistics therefore underestimate the true cross-document scope of these questions.
The \texttt{None}-answer subset has 218 questions (14.4\%) and tests whether systems can abstain when the documents do not contain the required support.

\subsection{Benchmark Reliability and Validation}
\label{sec:reliability}

Because \benchName{} uses LLM-generated candidates, reliability depends on human verification. We treat automatic filtering as candidate selection, not final validation. Of 3,550 synthetic candidates, 2,104 pass expert verification (59.3\%). We then remove lower-difficulty items that are too easy after verification, leaving \retainedQANum\ examples in the final release subset. Each retained example stores the question, final answer, answer format, reasoning type, expert-annotated evidence pages, evidence quotes or sentence-level snippets, and a typed verification rule. 
Human verification is a main part of the benchmark, not a light audit. In total, 194 human experts inspect retained candidates with full document access. They re-answer questions, mark the supporting pages and quotes, check whether the answer follows from the verification rule, flag ambiguity or missing support, and repair answer formats when the evidence is clear. This step is necessary because the automatic path-dependency check only works at the chapter or section level. Experts must convert that coarse signal into reliable page-level evidence. This is a main reason the benchmark is difficult to reproduce by prompting alone. Examples without clear evidence quotes and verification rules are removed.

\section{Experiments}
\label{sec:experiments}

\subsection{Evaluation Protocol}
\label{sec:eval_protocol}

\textbf{Input and pipeline settings.}
For inference, \emph{Img} denotes full-document page-image input, and \emph{OCR} denotes per-page OCR text capped at 80\% of the model context window so the model still has room to generate a stable answer. In both settings, the model receives the document representation and question, then returns a final answer without access to the expert-annotated evidence pages. Agent rows use PDF input; we use an LLM to extract the final answer from each agent response and then compute all metrics on the extracted answer.

\textbf{Models and metrics.}
Completed runs cover GPT-5.2, GPT-5.4, Claude Opus 4.6 \cite{anthropic2026claudeopus46}, Kimi-K2.5, DeepSeek-V3.2, Qwen3.5 9B/35B-A3B, and three RAG-based pipelines: MDocAgent, SimpleDoc, and DeepRead. Their nominal context windows are 1M tokens for GPT-5.4 and Claude Opus 4.6, 256K for GPT-5.2, Kimi-K2.5, and all Qwen3.5 variants, and 128K for DeepSeek-V3.2; with the 80\% OCR cap, the usable OCR budgets are about 800K, 205K, and 102K tokens, respectively. We report rule-based accuracy under each typed verification rule as the primary metric, plus token F1 and ANLS as secondary surface-form diagnostics. Failed, missing, or unparsable predictions count as incorrect.

\subsection{Main Results by Reasoning Type}
\label{sec:main_results}

\begin{table}[t]
\centering
\scriptsize
\setlength{\tabcolsep}{3pt}
\renewcommand{\arraystretch}{1.08}
\caption{\textbf{Main results on the human-verified \benchName{} subset.}
Per-reasoning-type accuracy, grouped into the three tiers of the taxonomy, followed by aggregate metrics.
\textbf{Img} (full document as page images), or \textbf{OCR} (per-page OCR text capped at 80\% of the model context window to reserve context for stable answer generation); agent frameworks operate directly on PDFs.
\textbf{T1 -- Core:} \textit{Cmp.} comparison, \textit{Ref.} reference chain.
\textbf{T2 -- Structural:} \textit{Rnk.} ranking, \textit{Cov.} coverage, \textit{Rec.} reconciliation.
\textbf{T3 -- Advanced:} \textit{SDiff} set difference, \textit{Unans} unanswerable, \textit{Tmp} temporal, \textit{Cmpl} compliance and rule-special-case checks, \textit{Ctf} counterfactual, \textit{Agg} aggregation, \textit{Con} consistency.
\textbf{Aggregate:} \textit{Acc.} overall accuracy, \textit{F1} token-level F1, \textit{ANLS} average normalised Levenshtein similarity. Per-column \bestcell{best} (green, bold) and \secondcell{second-best} (blue) values are highlighted; ties are all shaded.}
\label{tab:xldocbench-model-overall}
\resizebox{\linewidth}{!}{%
\begin{tabular}{l c cc ccc ccccccc ccc}
\toprule
 & & \multicolumn{2}{c}{T1 Core} & \multicolumn{3}{c}{T2 Structural} & \multicolumn{7}{c}{T3 Advanced} & \multicolumn{3}{c}{Aggregate} \\
\cmidrule(lr){3-4} \cmidrule(lr){5-7} \cmidrule(lr){8-14} \cmidrule(lr){15-17}
Model & In. & Cmp. & Ref. & Rnk. & Cov. & Rec. & SDiff & Unans & Tmp & Cmpl & Ctf & Agg & Con & Acc. & F1 & ANLS \\
\midrule
\multicolumn{17}{l}{\emph{Closed-source frontier}} \\
\multirow{2}{*}{GPT-5.2 (256K)} & Img & 9.8 & 13.7 & 10.8 & 12.6 & 18.3 & 13.6 & \bestcell{88.3} & 10.3 & 13.1 & 5.2 & 4.3 & 28.6 & 22.9 & 25.6 & 22.7 \\
                         & OCR & 19.1 & 28.2 & 17.6 & 25.8 & 28.7 & 25.4 & 83.6 & 13.8 & 26.2 & 13.8 & 4.3 & 19.0 & 30.9 & 33.2 & 28.1 \\
\multirow{2}{*}{GPT-5.4 (1M)} & Img & 28.9 & 26.9 & 20.5 & 18.7 & 23.5 & 20.3 & 44.9 & 24.1 & 26.2 & 22.4 & 23.4 & 28.6 & 26.9 & 29.6 & 25.9 \\
                         & OCR & 31.5 & 36.8 & 21.0 & \bestcell{30.8} & \secondcell{40.0} & \secondcell{33.9} & 70.6 & 36.2 & 32.8 & 31.0 & 27.7 & 23.8 & 37.3 & 39.5 & 34.1 \\
Claude Opus 4.6 (1M) & OCR & 38.7 & 36.8 & 29.0 & \secondcell{29.1} & 39.1 & 31.4 & 66.8 & 34.5 & 41.0 & 37.9 & 46.8 & \bestcell{47.6} & \secondcell{39.8} & \secondcell{40.5} & \secondcell{34.7} \\
\midrule
\multicolumn{17}{l}{\emph{Large open-source}} \\
\multirow{2}{*}{Kimi-K2.5 (256K)} & Img & 27.7 & 23.5 & 19.9 & 23.6 & 27.8 & 21.2 & 29.0 & \secondcell{37.9} & 23.0 & 32.8 & 25.5 & 9.5 & 25.4 & 27.0 & 22.5 \\
                           & OCR & 39.1 & \secondcell{39.3} & 30.1 & \secondcell{29.1} & 27.0 & 31.4 & 39.7 & 36.2 & \secondcell{44.3} & 41.4 & 42.6 & \secondcell{42.9} & 35.8 & 37.9 & 32.3 \\
DeepSeek-V3.2 (128K) & OCR & 23.0 & 26.5 & 17.0 & 19.2 & 21.7 & 24.6 & 78.5 & 17.2 & 18.0 & 15.5 & 23.4 & 23.8 & 29.6 & 32.8 & 29.0 \\
\midrule
\multicolumn{17}{l}{\emph{Small open-source (size series)}} \\
\multirow{2}{*}{Qwen3.5 9B (256K)} & Img & 17.9 & 18.4 & 18.8 & 12.1 & 9.6 & 13.6 & 38.8 & 25.9 & 21.3 & 12.1 & 23.4 & 23.8 & 19.8 & 22.6 & 20.1 \\
                               & OCR & 13.6 & 25.6 & 15.9 & 18.1 & 18.3 & 22.0 & 50.0 & 17.2 & 13.1 & 12.1 & 2.1 & 23.8 & 22.3 & 26.8 & 24.1 \\
\multirow{2}{*}{Qwen3.5 35B-A3B (256K)} & Img & 23.0 & 19.2 & 24.4 & 19.8 & 23.5 & 16.9 & 38.3 & 31.0 & 19.7 & 19.0 & 23.4 & \secondcell{42.9} & 24.2 & 17.5 & 11.6 \\
                               & OCR & 24.3 & 29.1 & 18.2 & 17.6 & 32.2 & 18.6 & \secondcell{84.1} & 19.0 & 27.9 & 17.2 & 31.9 & 23.8 & 32.0 & 35.8 & 32.5 \\
\midrule
\multicolumn{17}{l}{\emph{Agent framework}} \\
\multicolumn{17}{l}{\textbf{MDocAgent}} \\
\textit{ \quad + GPT-5.4 (1M)} & Agent & \secondcell{43.8} & 36.3 & \secondcell{33.5} & \bestcell{30.8} & 35.7 & 26.3 & 22.0 & \bestcell{46.6} & 39.3 & \secondcell{48.3} & \secondcell{48.9} & 38.1 & 35.0 & 35.8 & 28.4 \\
\textit{ \quad + Claude Opus 4.6 (1M)} & Agent & 42.1 & 37.6 & 29.5 & 24.7 & 36.5 & 25.4 & 21.5 & 44.8 & 39.3 & 43.1 & 40.4 & 42.9 & 33.2 & 34.9 & 28.6 \\
\textit{ \quad + GPT-5.2 (256K)} & Agent & 39.6 & 34.6 & 27.8 & 27.5 & 33.0 & 28.0 & 24.8 & 39.7 & 42.6 & 46.6 & 46.8 & 33.3 & 33.0 & 33.5 & 26.3 \\
\textit{ \quad + Kimi-K2.5 (256K)} & Agent & 39.6 & 33.3 & 27.3 & 23.6 & 33.9 & 25.4 & 21.5 & 39.7 & 34.4 & 46.6 & 38.3 & 33.3 & 31.1 & 33.6 & 27.5 \\
\textit{ \quad + DeepSeek-V3.2 (128K)} & Agent & 24.3 & 25.6 & 26.1 & 20.9 & 27.8 & 18.6 & 18.7 & 31.0 & 24.6 & 32.8 & 27.7 & 19.0 & 24.0 & 27.7 & 22.3 \\
\multicolumn{17}{l}{\textbf{SimpleDoc}} \\
\textit{ \quad + GPT-5.4 (1M)} & Agent & \bestcell{48.5} & \bestcell{41.0} & \bestcell{35.8} & 28.0 & \bestcell{41.7} & \bestcell{36.4} & 60.7 & \bestcell{46.6} & \bestcell{49.2} & \bestcell{55.2} & \bestcell{51.1} & \bestcell{47.6} & \bestcell{44.0} & \bestcell{43.7} & \bestcell{35.7} \\
\textit{ \quad + Claude Opus 4.6 (1M)} & Agent & 44.3 & 38.5 & 31.2 & 30.8 & 36.5 & 33.1 & 57.0 & 44.8 & 39.3 & 46.6 & 48.9 & 38.1 & 40.6 & 41.3 & 33.3 \\
\textit{ \quad + GPT-5.2 (256K)} & Agent & 37.9 & 33.8 & 33.0 & 25.8 & 34.8 & 26.3 & 68.2 & \secondcell{37.9} & 36.1 & 39.7 & 36.2 & 38.1 & 38.3 & 37.7 & 30.6 \\
\textit{ \quad + Kimi-K2.5 (256K)} & Agent & 34.9 & 33.3 & 23.9 & 23.1 & 31.3 & 26.3 & 55.6 & 34.5 & \secondcell{44.3} & 39.7 & 31.9 & 33.3 & 34.4 & 35.7 & 29.5 \\
\textit{ \quad + DeepSeek-V3.2 (128K)} & Agent & 32.8 & 34.2 & 22.7 & 20.3 & 33.9 & 28.0 & 76.2 & \secondcell{37.9} & 36.1 & 36.2 & 36.2 & 33.3 & 36.7 & 38.1 & 31.8 \\
\multicolumn{17}{l}{\textbf{DeepRead}} \\
\textit{ \quad + GPT-5.4 (1M)} & Agent & 35.3 & 37.6 & 27.8 & 29.1 & 38.3 & 32.2 & 27.6 & 29.3 & 36.1 & 25.9 & 29.8 & 33.3 & 32.2 & 33.2 & 26.8 \\
\textit{ \quad + Claude Opus 4.6 (1M)} & Agent & 26.0 & 24.4 & 23.9 & 25.3 & 27.8 & 17.0 & 32.7 & 22.4 & 31.1 & 36.2 & 25.5 & 33.3 & 26.3 & 28.3 & 23.1 \\
\textit{ \quad + GPT-5.2 (256K)} & Agent & 30.6 & 30.8 & 22.2 & 26.4 & 30.4 & 28.0 & 27.6 & 32.8 & 39.3 & 19.0 & 27.7 & 33.3 & 28.4 & 31.0 & 25.4 \\
\textit{ \quad + Kimi-K2.5 (256K)} & Agent & 27.7 & 24.4 & 22.7 & 18.7 & 20.0 & 21.2 & 14.0 & 24.1 & 32.8 & 27.6 & 31.9 & 23.8 & 22.6 & 25.0 & 21.5 \\
\textit{ \quad + DeepSeek-V3.2 (128K)} & Agent & 31.5 & 27.4 & 19.3 & 17.0 & 27.8 & 13.6 & 25.7 & 31.0 & 36.1 & 44.8 & 19.1 & 28.6 & 25.5 & 26.7 & 22.0 \\
\bottomrule
\end{tabular}%
}
\end{table}
Table~\ref{tab:xldocbench-model-overall} supports three observations.
\textbf{First, OCR is the strongest one-shot interface, but capacity alone is not the bottleneck.} OCR beats page-image input on every completed pure-model pair, with overall gains ranging from $+2.5$ pts (Qwen3.5-VL 9B: 19.8$\to$22.3) to $+10.4$ pts (GPT-5.4: 26.9$\to$37.3). The two 1M-token models nonetheless differ by 2.5 pts at equal usable OCR budget (Claude Opus 4.6 39.8\% vs.\ GPT-5.4 37.3\%), and 256K Kimi-K2.5 (35.8\%) remains close while 128K DeepSeek-V3.2 reaches only 29.6\% -- models must still locate the right pages, suppress irrelevant text, and apply the typed rule.
\textbf{Second, the hardest categories require tracking sets of evidence.} Best ranking, coverage, and set-difference accuracies stay at 35.8\%, 30.8\%, and 36.4\%, while SimpleDoc+GPT-5.4 reaches 55.2\% on counterfactual and 51.1\% on aggregation: once the right evidence is in context, models compute comparisons or counterfactuals reasonably well, but struggle to maintain a candidate set or check coverage. Claude Opus 4.6 OCR is the unusual one-shot exception, leading on Aggregation (46.8) and tying SimpleDoc+GPT-5.4 on Consistency (47.6) without retrieval. The trade-off shows on Unanswerable: Claude scores 66.8, below GPT-5.2 OCR (83.6), Qwen3.5-VL 35B-A3B OCR (84.1), and DeepSeek-V3.2 OCR (78.5) -- the strongest one-shot reader is also the most willing to attempt an answer when support is missing.
\textbf{Third, agent pipelines help only when retrieval produces usable context.} SimpleDoc+GPT-5.4 is the strongest system at 44.0\% ($+6.7$ over GPT-5.4 OCR), yet with the same backbone MDocAgent and DeepRead reach only 35.0\% and 32.2\%: the agent interface is not automatically better. Retrieval can also hurt abstention -- MDocAgent+GPT-5.4 scores 22.0\% on unanswerable while several one-shot OCR models exceed 78\% -- because partially relevant pages push the final model to answer when support is absent.

\begin{figure}[t]
    \centering
    \includegraphics[width=0.95\linewidth]{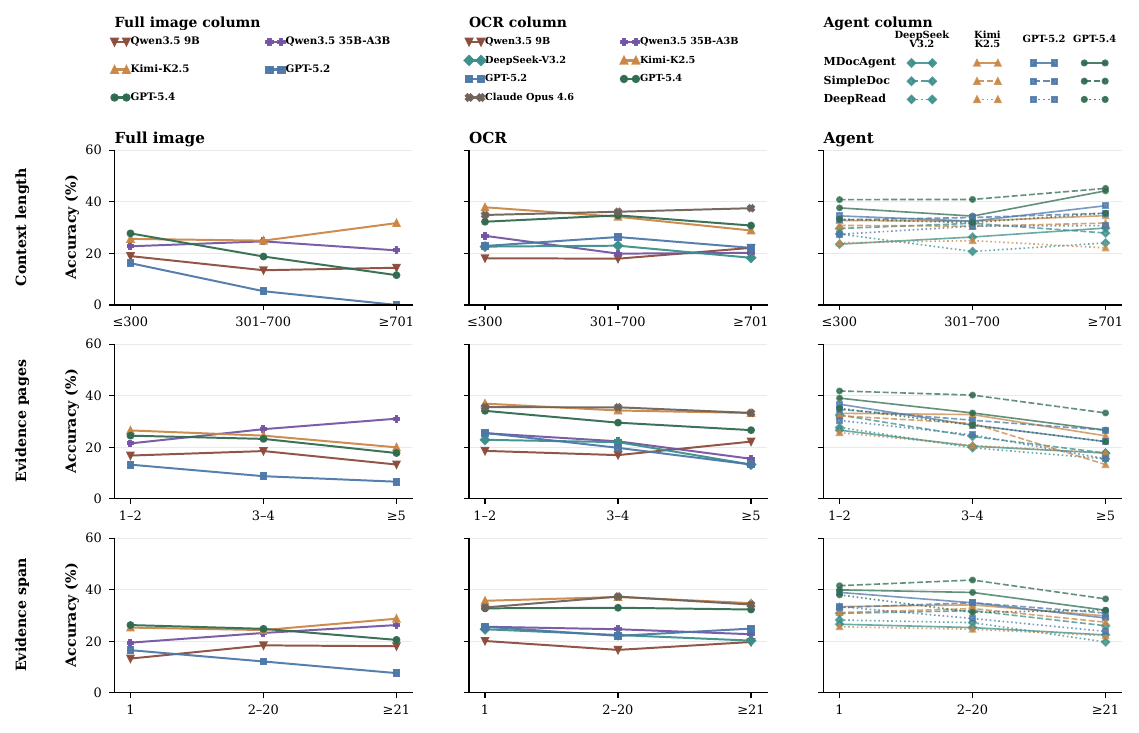}
    \caption{\textbf{Diagnostic performance breakdowns for all evaluated systems.}
    Columns separate full-image input, OCR input, and agent pipelines; rows report accuracy by context length, number of expert-annotated evidence pages, and evidence span.}
    \label{fig:exp_diagnostics}
\end{figure}
\subsection{Diagnostic Breakdowns}
\label{sec:diagnostic_breakdowns}

Aggregate accuracy does not explain why a system fails. Figure~\ref{fig:exp_diagnostics} decomposes accuracy by context length, evidence-page count, and evidence span -- three pressures that the overall score conflates. All direct readers degrade on longer contexts even though the 1M-token models (Claude Opus 4.6, GPT-5.4) and the 128K DeepSeek-V3.2 have very different windows, so the drop is not just truncation: models must still filter irrelevant pages and keep the right support active. OCR makes pages searchable but does not select evidence; agents pass compact context to a smaller backbone, but only when their search step finds every required page.

Table~\ref{tab:answer_type_breakdown_results} adds two checks. \textbf{Modality.} SimpleDoc+GPT-5.4 leads on text, tables, charts, and mixed modalities, while Kimi-K2.5 OCR and Claude Opus 4.6 OCR tie on image evidence -- the best overall pipeline is not the best visual reader. \textbf{Document scope.} Cross-document is not uniformly harder: GPT-5.4 OCR scores higher cross than single (43.03\% vs.\ 36.63\%), SimpleDoc+GPT-5.4 is balanced (43.03\% vs.\ 44.09\%), and Claude Opus 4.6 OCR is the only frontier one-shot reader with a small cross-document penalty ($-2.53$ vs.\ GPT-5.4 OCR's $+6.40$), suggesting its strength lies in deep single-document reading. Large negative gaps appear instead in unstable settings (DeepRead+DeepSeek-V3.2 $-14.30$, GPT-5.2 Img $-9.38$), so the harder step is selecting the right document and page, not the cross-document label itself.

\begin{figure}[t]
    \centering
    \includegraphics[width=0.98\linewidth]{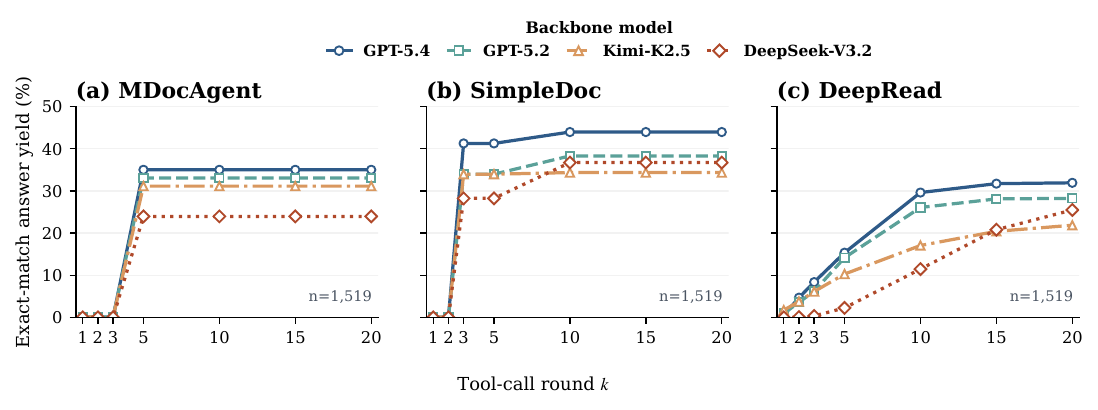}
    \caption{Agent answer yield versus tool-call rounds. Accuracy as the retrieval rounds increases.}
    \label{fig:fig_agent_answer_yield}
\end{figure}

\begin{figure}[t]
    \centering
    \includegraphics[width=0.98\linewidth]{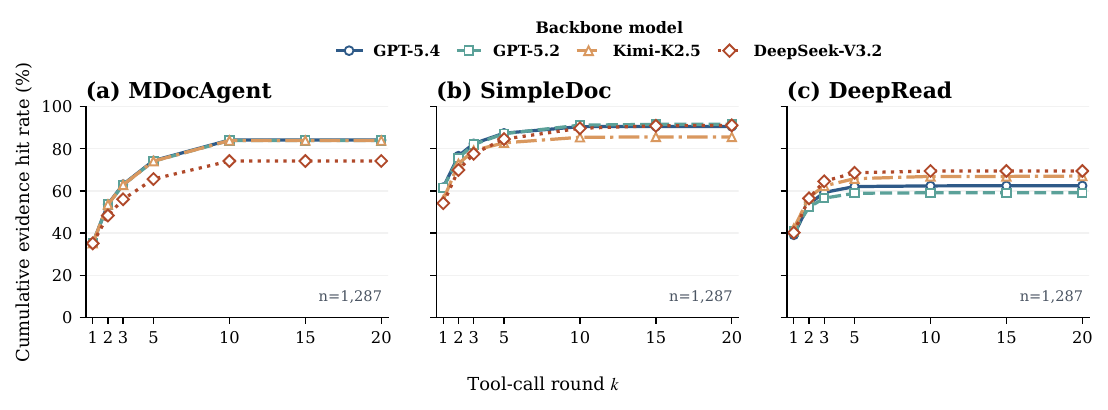}
    \caption{\textbf{Agent evidence-hit dynamics.} Evidence hit rate as the retrieved/read page rounds increases.}
    \label{fig:fig_agent_retrieval_dynamics}
\end{figure}

Figures~\ref{fig:fig_agent_answer_yield} and~\ref{fig:fig_agent_retrieval_dynamics} separate agent failures into retrieval and evidence-use errors. High evidence hit with low answer yield indicates a rule-following or answer-normalization problem; low evidence hit indicates a page-search or context-construction problem. The same final accuracy can therefore mask different failure modes.

\subsection{Discussion}
\label{sec:experiment_discussion}

The experiments show that window size is not the whole problem. The two 1M-token models (Claude Opus 4.6 and GPT-5.4) share the same usable OCR budget, yet Claude Opus 4.6 OCR -- itself a one-shot reader -- still leads GPT-5.4 OCR by 2.5 points and is in turn surpassed by the best agent pipeline. This indicates that capacity-driven gains and retrieval-driven gains are partially substitutable rather than additive. Some 256K and 128K models also improve when retrieval supplies compact evidence. OCR helps one-shot models by giving them searchable text, but it does not solve set tracking or abstention. Agent pipelines help when retrieval returns compact evidence, as in SimpleDoc, but can hurt when search misses a required page or supplies distracting context. These results show why long-document QA needs page-level evidence labels. Without them, retrieval failure and reasoning failure are hard to separate. The benchmark therefore reports reasoning-type accuracy, modality and scope breakdowns, and retrieval checks rather than a single leaderboard score.

\section{Conclusion}
\label{sec:conclusion}

We introduced \benchName{}, a fully human-verified benchmark for extra-long document understanding. It combines 1{,}519 questions over documents up to 2{,}303 pages with expert page-level evidence, typed verification rules, and a twelve-type reasoning taxonomy. Current long-context and agentic systems still struggle to find evidence, combine support, and satisfy explicit rules: the strongest pipeline reaches only 44.0\% overall, and the two 1M-token one-shot readers remain 4--7 points behind despite identical usable budgets. Per-type and per-modality breakdowns further show that set-tracking and abstention -- not raw context length -- are the binding constraints.

\bibliographystyle{plain}
\bibliography{references}

@article{ding2025survey,
  title={A Survey on MLLM-based Visually Rich Document Understanding: Methods, Challenges, and Emerging Trends},
  author={Ding, Yihao and Luo, Siwen and Dai, Yue and Jiang, Yanbei and Li, Zechuan and Martin, Geoffrey and Peng, Yifan},
  journal={arXiv preprint arXiv:2507.09861},
  year={2025}
}

@article{xia2024docgenome,
  title={Docgenome: An open large-scale scientific document benchmark for training and testing multi-modal large language models},
  author={Xia, Renqiu and Mao, Song and Yan, Xiangchao and Zhou, Hongbin and Zhang, Bo and Peng, Haoyang and Pi, Jiahao and Fu, Daocheng and Wu, Wenjie and Ye, Hancheng and others},
  journal={arXiv preprint arXiv:2406.11633},
  year={2024}
}

@inproceedings{bai2024longbench,
  title={Longbench: A bilingual, multitask benchmark for long context understanding},
  author={Bai, Yushi and Lv, Xin and Zhang, Jiajie and Lyu, Hongchang and Tang, Jiankai and Huang, Zhidian and Du, Zhengxiao and Liu, Xiao and Zeng, Aohan and Hou, Lei and others},
  booktitle={ACL},
  pages={3119--3137},
  year={2024}
}

@inproceedings{shaham2022scrolls,
  title={Scrolls: Standardized comparison over long language sequences},
  author={Shaham, Uri and Segal, Elad and Ivgi, Maor and Efrat, Avia and Yoran, Ori and Haviv, Adi and Gupta, Ankit and Xiong, Wenhan and Geva, Mor and Berant, Jonathan and others},
  booktitle={EMNLP},
  pages={12007--12021},
  year={2022}
}

@inproceedings{mathew2021docvqa,
  title={Docvqa: A dataset for vqa on document images},
  author={Mathew, Minesh and Karatzas, Dimosthenis and Jawahar, CV},
  booktitle={CVPR},
  pages={2200--2209},
  year={2021}
}

@inproceedings{masry2022chartqa,
  title={Chartqa: A benchmark for question answering about charts with visual and logical reasoning},
  author={Masry, Ahmed and Do, Xuan Long and Tan, Jia Qing and Joty, Shafiq and Hoque, Enamul},
  booktitle={ACL},
  pages={2263--2279},
  year={2022}
}

@inproceedings{mathew2022infographicvqa,
  title={Infographicvqa},
  author={Mathew, Minesh and Bagal, Viraj and Tito, Rub{\`e}n and Karatzas, Dimosthenis and Valveny, Ernest and Jawahar, CV},
  booktitle={WACV},
  pages={1697--1706},
  year={2022}
}

@inproceedings{zhu2021tat,
  title={TAT-QA: A question answering benchmark on a hybrid of tabular and textual content in finance},
  author={Zhu, Fengbin and Lei, Wenqiang and Huang, Youcheng and Wang, Chao and Zhang, Shuo and Lv, Jiancheng and Feng, Fuli and Chua, Tat-Seng},
  booktitle={IJNLP},
  pages={3277--3287},
  year={2021}
}

@article{tito2023hierarchical,
  title={Hierarchical multimodal transformers for multipage docvqa},
  author={Tito, Rub{\`e}n and Karatzas, Dimosthenis and Valveny, Ernest},
  journal={Pattern Recognition},
  volume={144},
  pages={109834},
  year={2023}
}

@article{liu2024visualwebbench,
  title={Visualwebbench: How far have multimodal llms evolved in web page understanding and grounding?},
  author={Liu, Junpeng and Song, Yifan and Lin, Bill Yuchen and Lam, Wai and Neubig, Graham and Li, Yuanzhi and Yue, Xiang},
  journal={arXiv preprint arXiv:2404.05955},
  year={2024}
}

@inproceedings{deng2025longdocurl,
  title={Longdocurl: a comprehensive multimodal long document benchmark integrating understanding, reasoning, and locating},
  author={Deng, Chao and Yuan, Jiale and Bu, Pi and Wang, Peijie and Li, Zhong-Zhi and Xu, Jian and Li, Xiao-Hui and Gao, Yuan and Song, Jun and Zheng, Bo and others},
  booktitle={ACL},
  pages={1135--1159},
  year={2025}
}

@inproceedings{van2023document,
  title={Document understanding dataset and evaluation (dude)},
  author={Van Landeghem, Jordy and Tito, Rub{\`e}n and Borchmann, {\L}ukasz and Pietruszka, Micha{\l} and Joziak, Pawel and Powalski, Rafal and Jurkiewicz, Dawid and Coustaty, Micka{\"e}l and Anckaert, Bertrand and Valveny, Ernest and others},
  booktitle={ICCV},
  pages={19528--19540},
  year={2023}
}

@inproceedings{tanaka2023slidevqa,
  title={Slidevqa: A dataset for document visual question answering on multiple images},
  author={Tanaka, Ryota and Nishida, Kyosuke and Nishida, Kosuke and Hasegawa, Taku and Saito, Itsumi and Saito, Kuniko},
  booktitle={AAAI},
  volume={37},
  number={11},
  pages={13636--13645},
  year={2023}
}

@article{ma2024mmlongbench,
  title={Mmlongbench-doc: Benchmarking long-context document understanding with visualizations},
  author={Ma, Yubo and Zang, Yuhang and Chen, Liangyu and Chen, Meiqi and Jiao, Yizhu and Li, Xinze and Lu, Xinyuan and Liu, Ziyu and Ma, Yan and Dong, Xiaoyi and others},
  journal={NeurIPS},
  volume={37},
  pages={95963--96010},
  year={2024}
}

@inproceedings{zou2025docbench,
  title={Docbench: A benchmark for evaluating llm-based document reading systems},
  author={Zou, Anni and Yu, Wenhao and Zhang, Hongming and Ma, Kaixin and Cai, Deng and Zhang, Zhuosheng and Zhao, Hai and Yu, Dong},
  booktitle={Proceedings of the 4th International Workshop on Knowledge-Augmented Methods for Natural Language Processing},
  pages={359--373},
  year={2025}
}

@inproceedings{ZhuLNWWFWLC26,
  author       = {Fengbin Zhu and
                  Ziyang Liu and
                  Xiang Yao Ng and
                  Haohui Wu and
                  Wenjie Wang and
                  Fuli Feng and
                  Chao Wang and
                  Huanbo Luan and
                  Tat{-}Seng Chua},
  title        = {MMDocBench: Benchmarking Large Vision-Language Models for Fine-Grained
                  Visual Document Understanding and Grounding},
  booktitle    = {MMM},
  pages        = {74--88},
  year         = {2026}
}

@inproceedings{ChiaCCSLAPB25,
  author       = {Yew Ken Chia and
                  Liying Cheng and
                  Hou Pong Chan and
                  Maojia Song and
                  Chaoqun Liu and
                  Mahani Aljunied and
                  Soujanya Poria and
                  Lidong Bing},
  title        = {M-LongDoc: {A} Benchmark For Multimodal Super-Long Document Understanding And {A} Retrieval-Aware Tuning Framework},
  booktitle    = {EMNLP},
  pages        = {9233--9250},
  year         = {2025}
}

@inproceedings{zhao2025finragbench,
  title={Finragbench-v: A benchmark for multimodal rag with visual citation in the financial domain},
  author={Zhao, Suifeng and Jin, Zhuoran and Li, Sujian and Gao, Jun},
  booktitle={EMNLP},
  pages={4215--4249},
  year={2025}
}

@misc{anthropic2026claudeopus46,
  author       = {{Anthropic}},
  title        = {Introducing Claude Opus 4.6},
  year         = {2026},
  howpublished = {\url{https://www.anthropic.com/news/claude-opus-4-6}},
}

@misc{Gemini3,
  author       = {{Google}},
  title        = {Gemini-3},
  year         = {2026},
  howpublished = {\url{https://aistudio.google.com/models/gemini-3/}},
}

@misc{Qwen35,
  author = {{Qwen Team}},
  title = {Qwen3.5: Towards Native Multimodal Agents},
  year = {2026},
  howpublished = {\url{https://qwen.ai/blog?id=qwen3.5}},
}

@inproceedings{jain2025simpledoc,
  title={SimpleDoc: Multi-Modal Document Understanding with Dual-Cue Page Retrieval and Iterative Refinement},
  author={Jain, Chelsi and Wu, Yiran and Zeng, Yifan and Liu, Jiale and Dai, Shengyu and Shao, Zhenwen and Wu, Qingyun and Wang, Huazheng},
  booktitle={Proceedings of the 2025 Conference on Empirical Methods in Natural Language Processing},
  pages={28398--28415},
  year={2025}
}

@article{han2025mdocagent,
  title={Mdocagent: A multi-modal multi-agent framework for document understanding},
  author={Han, Siwei and Xia, Peng and Zhang, Ruiyi and Sun, Tong and Li, Yun and Zhu, Hongtu and Yao, Huaxiu},
  journal={arXiv preprint arXiv:2503.13964},
  year={2025}
}

@article{li2026deepread,
  title={DeepRead: Document Structure-Aware Reasoning to Enhance Agentic Search},
  author={Li, Zhanli and Tian, Huiwen and Luo, Lvzhou and Cao, Yixuan and Luo, Ping},
  journal={arXiv preprint arXiv:2602.05014},
  year={2026}
}

@inproceedings{xu2020layoutlm,
  title={Layoutlm: Pre-training of text and layout for document image understanding},
  author={Xu, Yiheng and Li, Minghao and Cui, Lei and Huang, Shaohan and Wei, Furu and Zhou, Ming},
  booktitle={KDD},
  pages={1192--1200},
  year={2020}
}

@inproceedings{xu2021layoutlmv2,
  title={Layoutlmv2: Multi-modal pre-training for visually-rich document understanding},
  author={Xu, Yang and Xu, Yiheng and Lv, Tengchao and Cui, Lei and Wei, Furu and Wang, Guoxin and Lu, Yijuan and Florencio, Dinei and Zhang, Cha and Che, Wanxiang and others},
  booktitle={IJCNLP},
  pages={2579--2591},
  year={2021}
}

@inproceedings{huang2022layoutlmv3,
  title={Layoutlmv3: Pre-training for document ai with unified text and image masking},
  author={Huang, Yupan and Lv, Tengchao and Cui, Lei and Lu, Yutong and Wei, Furu},
  booktitle={ACM MM},
  pages={4083--4091},
  year={2022}
}

@inproceedings{wang2022lilt,
  title={Lilt: A simple yet effective language-independent layout transformer for structured document understanding},
  author={Wang, Jiapeng and Jin, Lianwen and Ding, Kai},
  booktitle={ACL},
  pages={7747--7757},
  year={2022}
}

@inproceedings{appalaraju2021docformer,
  title={Docformer: End-to-end transformer for document understanding},
  author={Appalaraju, Srikar and Jasani, Bhavan and Kota, Bhargava Urala and Xie, Yusheng and Manmatha, R},
  booktitle={ICCV},
  pages={993--1003},
  year={2021}
}

@article{kim2021donut,
  title={Donut: Document understanding transformer without ocr},
  author={Kim, Geewook and Hong, Teakgyu and Yim, Moonbin and Park, Jinyoung and Yim, Jinyeong and Hwang, Wonseok and Yun, Sangdoo and Han, Dongyoon and Park, Seunghyun},
  journal={arXiv preprint arXiv:2111.15664},
  volume={7},
  number={15},
  pages={2},
  year={2021}
}

@article{ye2023mplug,
  title={mplug-docowl: Modularized multimodal large language model for document understanding},
  author={Ye, Jiabo and Hu, Anwen and Xu, Haiyang and Ye, Qinghao and Yan, Ming and Dan, Yuhao and Zhao, Chenlin and Xu, Guohai and Li, Chenliang and Tian, Junfeng and others},
  journal={arXiv preprint arXiv:2307.02499},
  year={2023}
}

@inproceedings{hu2025mplug,
  title={mplug-docowl2: High-resolution compressing for ocr-free multi-page document understanding},
  author={Hu, Anwen and Xu, Haiyang and Zhang, Liang and Ye, Jiabo and Yan, Ming and Zhang, Ji and Jin, Qin and Huang, Fei and Zhou, Jingren},
  booktitle={ACL},
  pages={5817--5834},
  year={2025}
}

@article{liu2026textmonkey,
  title={Textmonkey: An ocr-free large multimodal model for understanding document},
  author={Liu, Yuliang and Yang, Biao and Liu, Qiang and Li, Zhang and Ma, Zhiyin and Zhang, Shuo and Bai, Xiang},
  journal={IEEE Transactions on Pattern Analysis and Machine Intelligence},
  year={2026},
}

@article{wang2025internvl3,
  title={Internvl3. 5: Advancing open-source multimodal models in versatility, reasoning, and efficiency},
  author={Wang, Weiyun and Gao, Zhangwei and Gu, Lixin and Pu, Hengjun and Cui, Long and Wei, Xingguang and Liu, Zhaoyang and Jing, Linglin and Ye, Shenglong and Shao, Jie and others},
  journal={arXiv preprint arXiv:2508.18265},
  year={2025}
}

@article{agrawal2024pixtral,
  title={Pixtral 12B},
  author={Agrawal, Pravesh and Antoniak, Szymon and Hanna, Emma Bou and Bout, Baptiste and Chaplot, Devendra and Chudnovsky, Jessica and Costa, Diogo and De Monicault, Baudouin and Garg, Saurabh and Gervet, Theophile and others},
  journal={arXiv preprint arXiv:2410.07073},
  year={2024}
}

@article{faysse2024colpali,
  title={Colpali: Efficient document retrieval with vision language models},
  author={Faysse, Manuel and Sibille, Hugues and Wu, Tony and Omrani, Bilel and Viaud, Gautier and Hudelot, C{\'e}line and Colombo, Pierre},
  journal={arXiv preprint arXiv:2407.01449},
  year={2024}
}

@inproceedings{ma2025towards,
  title={Towards storage-efficient visual document retrieval: An empirical study on reducing patch-level embeddings},
  author={Ma, Yubo and Li, Jinsong and Zang, Yuhang and Wu, Xiaobao and Dong, Xiaoyi and Zhang, Pan and Cao, Yuhang and Duan, Haodong and Wang, Jiaqi and Cao, Yixin and others},
  booktitle={ACL},
  pages={19568--19580},
  year={2025}
}

@article{yu2024visrag,
  title={Visrag: Vision-based retrieval-augmented generation on multi-modality documents},
  author={Yu, Shi and Tang, Chaoyue and Xu, Bokai and Cui, Junbo and Ran, Junhao and Yan, Yukun and Liu, Zhenghao and Wang, Shuo and Han, Xu and Liu, Zhiyuan and others},
  journal={arXiv preprint arXiv:2410.10594},
  year={2024}
}

@article{cho2024m3docrag,
  title={M3docrag: Multi-modal retrieval is what you need for multi-page multi-document understanding},
  author={Cho, Jaemin and Mahata, Debanjan and Irsoy, Ozan and He, Yujie and Bansal, Mohit},
  journal={arXiv preprint arXiv:2411.04952},
  year={2024}
}

@inproceedings{tanaka2025vdocrag,
  title={Vdocrag: Retrieval-augmented generation over visually-rich documents},
  author={Tanaka, Ryota and Iki, Taichi and Hasegawa, Taku and Nishida, Kyosuke and Saito, Kuniko and Suzuki, Jun},
  booktitle={CVPR},
  pages={24827--24837},
  year={2025}
}

\newpage
\appendix
\begin{table}[t]
\centering
\scriptsize
\setlength{\tabcolsep}{3pt}
\renewcommand{\arraystretch}{1.05}
\caption{Accuracy by document domain for all evaluated systems. Bold pipeline names are group headers; rows underneath report pipeline--backbone combinations.}
\label{tab:domain_breakdown_results}
\resizebox{\linewidth}{!}{%
\begin{tabular}{lrrrrrr}
\toprule
Model/Input & Legal reg. & Finance bus. & Technical eng. & Medical clin. & Scientific acad. & Narrative lit. \\
\midrule
\multicolumn{7}{l}{\emph{Closed-source frontier}} \\
GPT-5.2 Img & 24.8 & 21.0 & 24.8 & 22.1 & 21.6 & 23.6 \\
GPT-5.2 OCR & 28.9 & 26.7 & 32.1 & 33.7 & 33.5 & 33.7 \\
GPT-5.4 Img & 29.6 & 29.7 & 24.5 & 24.1 & 25.2 & 27.0 \\
GPT-5.4 OCR & 34.4 & 37.5 & 36.1 & 38.1 & 40.8 & 39.3 \\
Claude Opus 4.6 OCR & 38.9 & 38.1 & 39.4 & 41.2 & 42.7 & 39.3 \\
\midrule
\multicolumn{7}{l}{\emph{Large open-source}} \\
Kimi-K2.5 Img & 25.4 & 28.8 & 24.8 & 22.1 & 26.1 & 23.6 \\
Kimi-K2.5 OCR & 33.1 & 34.5 & 35.8 & 36.4 & 39.4 & 39.3 \\
DeepSeek-V3.2 OCR & 28.6 & 26.4 & 35.4 & 29.3 & 28.9 & 29.2 \\
\midrule
\multicolumn{7}{l}{\emph{Small open-source (size series)}} \\
Qwen3.5 9B Img & 19.9 & 18.9 & 20.1 & 22.1 & 18.8 & 16.9 \\
Qwen3.5 9B OCR & 19.0 & 21.0 & 24.1 & 24.8 & 21.1 & 27.0 \\
Qwen3.5 35B-A3B Img & 29.6 & 21.6 & 27.0 & 23.5 & 20.2 & 19.1 \\
Qwen3.5 35B-A3B OCR & 31.5 & 30.0 & 30.7 & 33.3 & 35.8 & 31.5 \\
\midrule
\multicolumn{7}{l}{\emph{Agent pipelines (PDF input)}} \\
\textbf{MDocAgent} \\
\quad + GPT-5.4 & 35.4 & 35.1 & 37.6 & 33.7 & 35.3 & 29.2 \\
\quad + Claude Opus 4.6 & 32.5 & 34.2 & 38.0 & 31.6 & 34.4 & 20.2 \\
\quad + GPT-5.2 & 35.0 & 30.0 & 36.5 & 28.9 & 38.1 & 28.1 \\
\quad + Kimi-K2.5 & 33.4 & 34.8 & 32.1 & 26.2 & 33.5 & 16.9 \\
\quad + DeepSeek-V3.2 & 22.5 & 24.3 & 30.7 & 19.7 & 23.9 & 21.3 \\
\addlinespace[2pt]
\textbf{SimpleDoc}  \\
\quad + GPT-5.4 & 40.2 & 42.6 & 44.9 & 42.5 & 52.3 & 43.8 \\
\quad + Claude Opus 4.6 & 39.2 & 36.9 & 43.4 & 39.1 & 44.5 & 44.9 \\
\quad + GPT-5.2 & 38.9 & 34.8 & 43.1 & 36.7 & 39.9 & 36.0 \\
\quad + Kimi-K2.5 & 31.8 & 35.7 & 35.0 & 34.4 & 33.5 & 38.2 \\
\quad + DeepSeek-V3.2 & 36.0 & 32.1 & 41.6 & 34.0 & 40.4 & 41.6 \\
\addlinespace[2pt]
\textbf{DeepRead}  \\
\quad + GPT-5.4 & 31.8 & 30.9 & 32.8 & 32.3 & 34.9 & 29.2 \\
\quad + Claude Opus 4.6 & 26.7 & 25.5 & 29.9 & 25.2 & 26.6 & 20.2 \\
\quad + GPT-5.2 & 27.3 & 26.1 & 32.8 & 26.9 & 28.9 & 31.5 \\
\quad + Kimi-K2.5 & 27.0 & 17.4 & 23.4 & 22.1 & 23.4 & 24.7 \\
\quad + DeepSeek-V3.2 & 22.5 & 24.6 & 26.3 & 25.2 & 30.7 & 24.7 \\
\bottomrule
\end{tabular}}
\end{table}

\begin{table}[!t]
\centering
\scriptsize
\setlength{\tabcolsep}{3pt}
\renewcommand{\arraystretch}{1.0}
\caption{Accuracy by evidence modality and document scope for all evaluated systems. Gap is Cross--Single, so negative values indicate a cross-document penalty. Text-combination modalities are assigned to the non-text modality (e.g., Table+Text $\rightarrow$ Table); combinations of multiple non-text modalities are grouped as Mix.}
\label{tab:answer_type_breakdown_results}
\resizebox{0.9\linewidth}{!}{%
\begin{tabular}{lrrrrrrrr}
\toprule
& \multicolumn{5}{c}{Evidence modality} & \multicolumn{3}{c}{Document scope} \\
\cmidrule(lr){2-6}\cmidrule(lr){7-9}
Model/Input & Text & Table & Chart & Image & Mix & Single & Cross & Gap \\
\midrule
\multicolumn{9}{l}{\emph{Closed-source frontier}} \\
GPT-5.2 Img & 24.3 & 20.0 & 21.9 & 18.2 & 20.0 & 23.93 & 14.55 & -9.38 \\
GPT-5.2 OCR & 32.4 & 26.4 & 31.6 & 36.4 & 28.3 & 31.39 & 26.67 & -4.72 \\
GPT-5.4 Img & 28.0 & 22.7 & 28.4 & 27.3 & 26.7 & 27.40 & 22.42 & -4.98 \\
GPT-5.4 OCR & 38.3 & 37.3 & 34.8 & 36.4 & 28.3 & 36.63 & 43.03 & 6.40 \\
Claude Opus 4.6 OCR & 39.5 & 38.8 & 43.9 & 45.5 & 40.0 & 40.10 & 37.58 & -2.53 \\
\midrule
\multicolumn{9}{l}{\emph{Large open-source}} \\
Kimi-K2.5 Img & 28.0 & 20.3 & 21.3 & 36.4 & 20.0 & 24.00 & 36.97 & 12.97 \\
Kimi-K2.5 OCR & 36.9 & 33.3 & 34.2 & 45.5 & 35.0 & 36.34 & 31.52 & -4.82 \\
DeepSeek-V3.2 OCR & 30.4 & 27.6 & 27.7 & 27.3 & 31.7 & 29.99 & 26.06 & -3.92 \\
\midrule
\multicolumn{9}{l}{\emph{Small open-source (size series)}} \\
Qwen3.5 9B Img & 20.8 & 17.9 & 20.6 & 27.3 & 11.7 & 19.57 & 21.82 & 2.25 \\
Qwen3.5 9B OCR & 23.3 & 21.2 & 18.7 & 18.2 & 21.7 & 21.57 & 27.88 & 6.31 \\
Qwen3.5 35B-A3B Img & 26.5 & 20.3 & 22.6 & 27.3 & 13.3 & 24.08 & 25.45 & 1.38 \\
Qwen3.5 35B-A3B OCR & 33.5 & 28.5 & 29.7 & 18.2 & 35.0 & 32.87 & 24.85 & -8.02 \\
\midrule
\multicolumn{9}{l}{\emph{Agent pipelines (PDF input)}} \\
\textbf{MDocAgent}\\
\quad + GPT-5.4 & 35.2 & 34.2 & 38.7 & 18.2 & 30.0 & 34.49 & 39.39 & 4.90 \\
\quad + Claude Opus 4.6 & 33.3 & 30.6 & 43.2 & 18.2 & 23.3 & 34.27 & 24.85 & -9.42 \\
\quad + GPT-5.2 & 33.6 & 29.7 & 40.6 & 9.1 & 26.7 & 32.72 & 35.76 & 3.04 \\
\quad + Kimi-K2.5 & 31.2 & 30.0 & 34.8 & 27.3 & 28.3 & 30.95 & 32.73 & 1.78 \\
\quad + DeepSeek-V3.2 & 25.2 & 21.5 & 24.5 & 18.2 & 16.7 & 23.56 & 27.27 & 3.71 \\
\addlinespace[2pt]
\textbf{SimpleDoc}  \\
\quad + GPT-5.4 & 44.7 & 43.0 & 45.8 & 9.1 & 40.0 & 44.09 & 43.03 & -1.06 \\
\quad + Claude Opus 4.6 & 41.5 & 39.7 & 42.6 & 18.2 & 28.3 & 40.32 & 42.42 & 2.10 \\
\quad + GPT-5.2 & 38.3 & 39.1 & 41.9 & 9.1 & 30.0 & 38.11 & 40.00 & 1.89 \\
\quad + Kimi-K2.5 & 36.8 & 28.8 & 34.2 & 18.2 & 30.0 & 33.90 & 38.18 & 4.28 \\
\quad + DeepSeek-V3.2 & 37.6 & 34.2 & 38.1 & 18.2 & 36.7 & 36.63 & 37.58 & 0.94 \\
\addlinespace[2pt]
\textbf{DeepRead} &  \\
\quad + GPT-5.4 & 32.7 & 32.7 & 29.7 & 18.2 & 30.0 & 32.27 & 31.52 & -0.76 \\
\quad + Claude Opus 4.6 & 26.3 & 27.6 & 25.8 & 18.2 & 23.3 & 27.03 & 20.61 & -6.42 \\
\quad + GPT-5.2 & 29.7 & 26.4 & 26.5 & 36.4 & 23.3 & 28.73 & 26.06 & -2.67 \\
\quad + Kimi-K2.5 & 22.4 & 24.5 & 19.4 & 18.2 & 25.0 & 23.04 & 19.39 & -3.65 \\
\quad + DeepSeek-V3.2 & 24.4 & 27.0 & 24.5 & 27.3 & 36.7 & 27.03 & 12.73 & -14.30 \\
\bottomrule
\end{tabular}}
\end{table}

\section{Construction Pipeline Details}
\label{app:pipeline_details}

This appendix gives implementation details for the construction pipeline in \S\ref{sec:pipeline}. Each document is parsed into page images, OCR text, Markdown, tables, and figure metadata. Headings, tables of contents, layout cues, and fallback page groups define a hierarchical tree; for document series, per-document trees are attached to a shared root. Candidate generation uses branch summaries, domain quotas, and reasoning-type quotas. We keep candidates that pass leave-one-branch-out checks at the chapter or section level, iterative refinement checks, grounding checks, and artifact filters. The released page-level evidence does not come from this coarse automatic stage. It is annotated and verified by 194 human experts with full document access.

The filters target five failure modes: malformed answer schema, meaningless arithmetic such as page-number subtraction, failed branch-level path-dependency, answerability without document evidence, and residual references to internal tree nodes. The final released subset contains 1,519 examples from 3,550 generated candidates. The no-context filter is the largest single rejection source. It gives the judge only the question and no document content; if the answer is still inferable from world knowledge, metadata, or wording leakage, the candidate is rejected. After expert verification, we also remove lower-difficulty items that pass support checks but do not sufficiently test long-context evidence localization or multi-step reasoning. This produces the final 1,519-example release subset.

\section{Diagnostic Breakdown Tables}
\label{app:diagnostic_breakdown_tables}

Tables~\ref{tab:domain_breakdown_results} and~\ref{tab:answer_type_breakdown_results} provide breakdowns by document domain, evidence modality, and document scope for the evaluated systems.

\section{Evaluation Protocol Details}
\label{app:eval_protocol}

Each example includes an answer type and a typed verification rule. Integer answers require exact normalized match. Float answers specify units, rounding, and tolerance. String answers include aliases and are evaluated with normalized alias/rule matching for accuracy, with token F1 and ANLS as secondary surface-form metrics. Unanswerable examples use \texttt{None} and include a rationale describing which required condition is absent from the documents. These rules connect construction-time grounding to evaluation-time scoring and reduce the need for unconstrained free-form judging.

\section{Dataset Card}
\label{app:dataset_card}

The released dataset card will document the benchmark motivation, composition, source documents, preprocessing, annotation process, validation process, intended use, out-of-scope use, licensing, maintenance plan, and known limitations. The benchmark is intended for evaluating document-understanding systems, not for making financial, medical, legal, or regulatory decisions.

\section{Broader Impacts and Limitations}
\label{sec:broader_impacts_limitations}

The benchmark supports research on reliable document understanding, evidence-based evaluation, and long-context system design. It can help reveal unsupported or poorly supported document answers before such systems are used in professional settings. At the same time, high benchmark performance does not show that a model is safe for legal, medical, financial, regulatory, or other high-stakes work. The dataset uses public professional documents, but evaluated systems may still give misleading answers, fail to abstain, or cite incomplete evidence. Users should treat \benchName{} as a research benchmark and diagnostic tool, not as a replacement for domain experts or formal review. Limitations include possible style bias from LLM-assisted candidate generation, finite human verification capacity, public-document exposure in model pretraining, English and public-source coverage limits, residual PDF parsing or OCR errors, and the smaller size of the cross-document subset compared with the full benchmark.

\section{Declaration of LLM Usage}
\label{sec:llm_usage}

We used LLMs in the construction pipeline to propose candidate questions, provisional answers, candidate evidence, verification rules, and automatic filtering or judging signals, as described in Appendix~\ref{app:prompts_and_eval}. These LLM-generated outputs were provisional: LLMs were not used to replace human verification, determine the final benchmark labels, or make final decisions about the reported results. All released answers, evidence annotations, reasoning labels, statistics, experimental conclusions, and paper edits were reviewed by the authors.

\section{Failure Case Studies}
\label{app:failure_case_studies}

\begin{figure}[p]
    \centering
    \includegraphics[width=\linewidth,height=0.92\textheight,keepaspectratio]{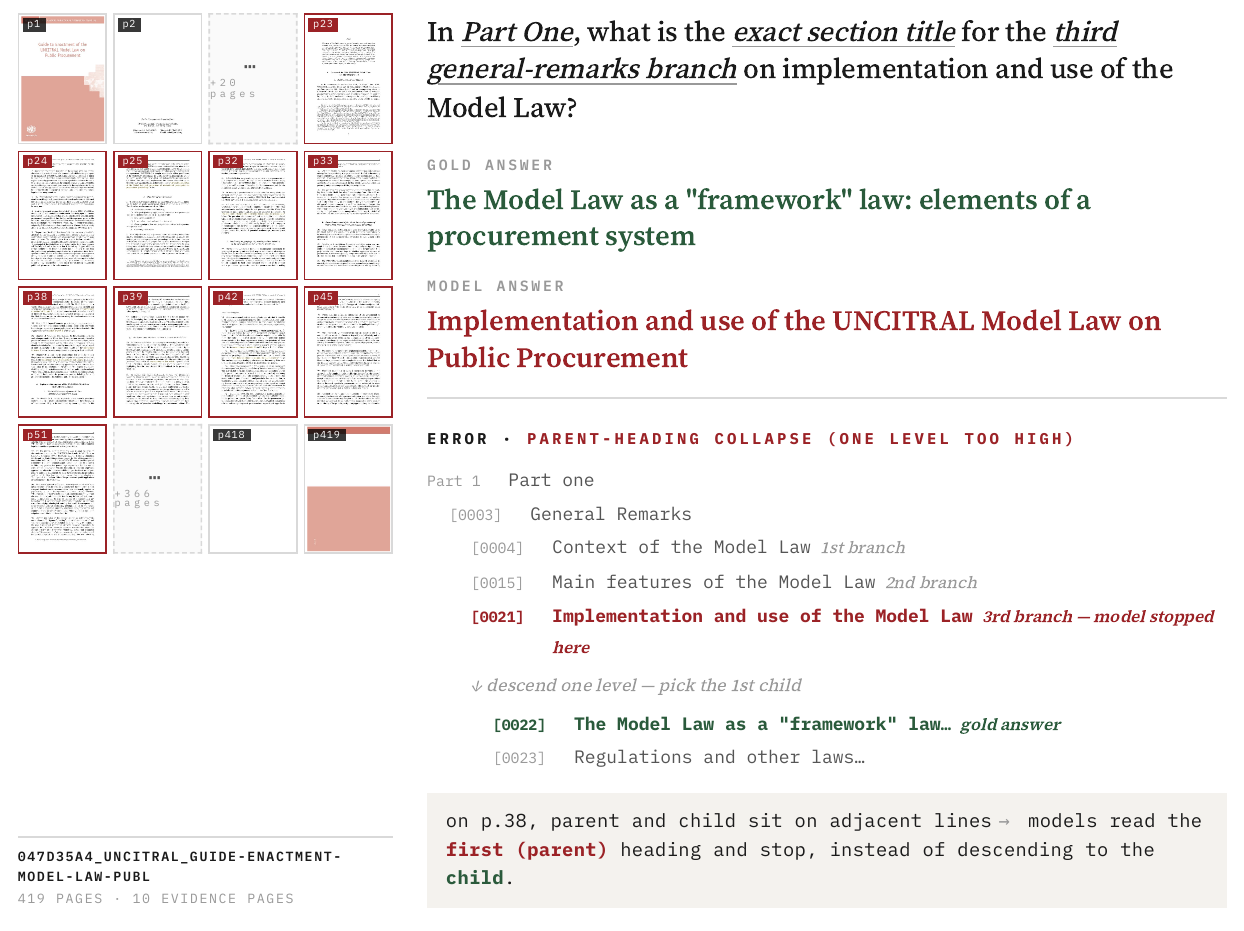}
    \caption{Failure case study 1. The example illustrates a system failure on an evidence-grounded long-document question. The model response does not satisfy the verified answer rule, showing how long-context QA can fail even when the question appears answerable from the document.}
    \label{fig:failed_case_1}
\end{figure}

\begin{figure}[p]
    \centering
    \includegraphics[width=\linewidth,height=0.92\textheight,keepaspectratio]{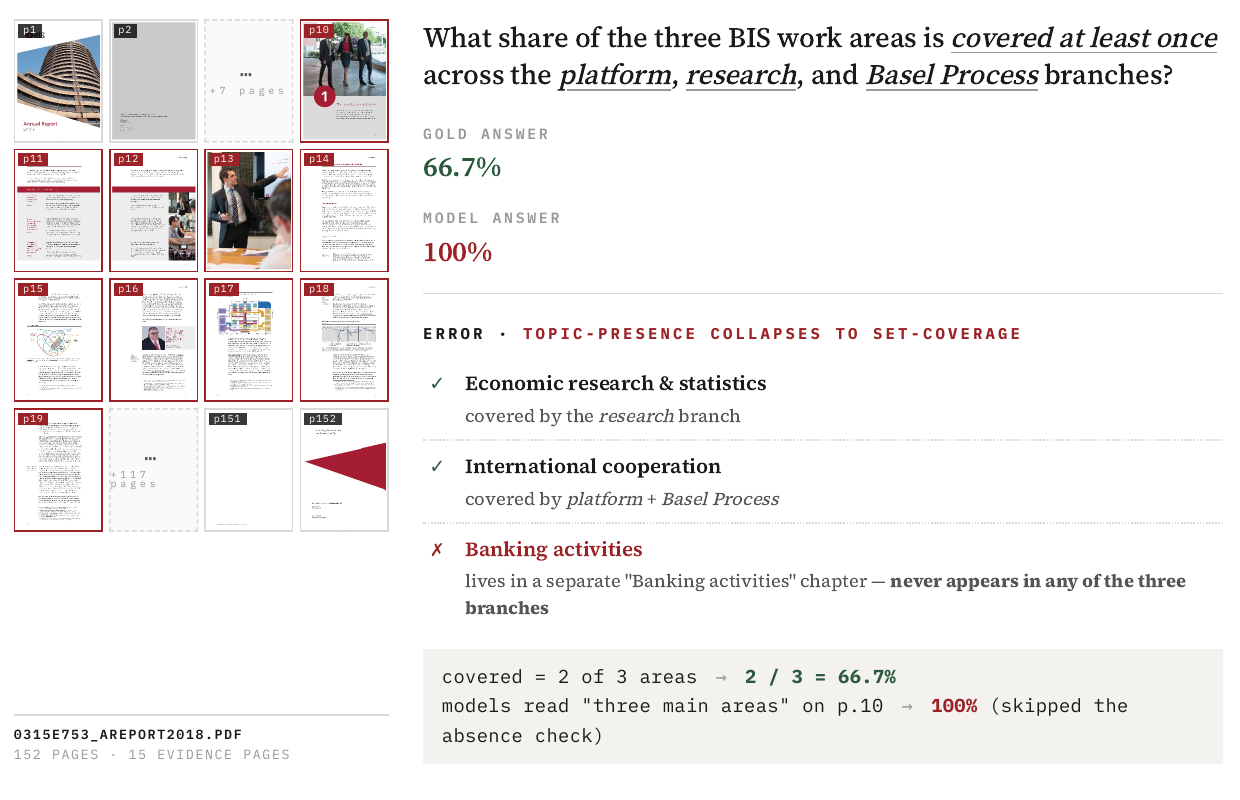}
    \caption{Failure case study 2. The final response is inconsistent with the expert-verified evidence. Such cases motivate reporting evidence-use checks in addition to final-answer accuracy.}
    \label{fig:failed_case_2}
\end{figure}

Figures~\ref{fig:failed_case_1} and~\ref{fig:failed_case_2} show representative failed cases from the evaluated systems. These examples appear before the reasoning-type case studies and show the main error modes discussed in \S\ref{sec:diagnostic_breakdowns}. In both cases, the model output is not merely a surface-form mismatch: the failure reflects a breakdown in evidence search, evidence use, or rule-following answer generation. The examples also explain why \benchName{} records page-level evidence and typed verification rules. Without these annotations, it is difficult to tell whether the system failed because it missed the supporting page, used the wrong evidence, or answered without satisfying the required condition.

\clearpage
\section{Synthesis Prompts and Evaluation Algorithms}
\label{app:prompts_and_eval}

This section summarizes the prompts used by the data-synthesis pipeline and the deterministic scoring procedure used in our experiments. The boxes describe each stage's role, constraints, and output format rather than the full implementation strings.

\subsection{Data-Synthesis Prompt Structure}
\label{app:synthesis_prompts}

\begin{tcolorbox}[promptbox,title=\textbf{Four-stage synthesis chain}]
\small
\textbf{Exploration} proposes root-level questions from the document tree. The prompt asks the model to act as a benchmark designer rather than a summarizer: it should identify unresolved issues that require descending into lower sections, prefer verifiable closed-form answers, and assign a reasoning type. The pipeline rejects questions that can be answered from the root summary alone.\\[2pt]
\textbf{Refinement} rewrites the parent question using child-node summaries without changing the topic. It narrows the answer space, selects the smallest sufficient set of next nodes, and keeps cross-branch or cross-document dependence when the item is designed to span multiple branches.\\[2pt]
\textbf{Evidence decomposition} turns the refined question into atomic support checks. Each sub-question must isolate one concrete support point, preferably from a distinct page or section, and returns a compact answer plus a verification rule.\\[2pt]
\textbf{Final decision} writes the benchmark-ready QA item. The prompt enforces answer-format constraints, rejects trivial arithmetic and metadata-derived answers, preserves the intended reasoning type, and outputs a provisional question, answer, candidate evidence pages, and typed verification rule for subsequent human verification.
\end{tcolorbox}

\begin{tcolorbox}[promptbox,title=\textbf{Quality constraints used across prompts}]
\small
\textbf{Path dependency} requires that a candidate cannot be answered from high-level summaries alone. The automatic check operates over chapter or section branches; final page-level evidence is added by human experts.\\[2pt]
\textbf{Cross-branch necessity} requires that removing any targeted branch weakens or breaks the question. For multi-document trees, the prompt forces evidence from at least two PDFs and uses document-prefixed node IDs.\\[2pt]
\textbf{Answer discipline} bans boolean answers, counting-only questions, page-number arithmetic, ID arithmetic, and vague multi-part questions. Numeric answers must have meaningful units or domains, entity answers must be short, and single-choice options must be mutually exclusive.\\[2pt]
\textbf{Artifact filtering} rejects candidates answerable from world knowledge, document metadata, or wording leakage. It also rejects questions whose apparent difficulty comes from formatting noise rather than document understanding.
\end{tcolorbox}

\begin{tcolorbox}[promptbox,title=\textbf{Editing and judge prompts}]
\small
\textbf{Packaging editor} shortens the final QA without changing its meaning or supported answer. It removes unnecessary section numbering, keeps at least one phrase that helps retrieval, and normalizes answer surfaces.\\[2pt]
\textbf{Page-stage repair editor} is called when convergence checks fail. It restores lost constraints, repairs drift from the refined question, and re-inserts important fields such as criterion, threshold, comparison target, or mismatch.\\[2pt]
\textbf{Path-dependency judge} decides whether the candidate still requires hierarchical navigation. A high score means the answer needs evidence from multiple pages, sections, or tables rather than a single local span.\\[2pt]
\textbf{State-assessment judge} scores answerability, confidence, and benchmarkability for an in-flight exploration state. It also flags summary-shortcut risk and ambiguity risk before the candidate reaches human verification.
\end{tcolorbox}

\subsection{Evaluation Code Sketch}
\label{app:evaluation_pseudocode}

Our reported metrics are computed by deterministic scripts. Direct baselines are instructed to put the final answer in \texttt{<answer>} tags; for agent baselines, we first extract a concise final answer from the agent response and then run the same scoring code. The headline accuracy does not use an LLM judge. The simplified Python-style sketch below shows the main logic.

\begin{lstlisting}[style=pyappendix,caption={Python-style sketch of the deterministic evaluator.},label={lst:evaluation_sketch}]
def evaluate_prediction(response, gold, answer_type):
    pred = extract_answer(response)          # <answer>...</answer>, prefix, or final line
    pred_n = normalize(pred)
    gold_n = normalize(gold)

    if answer_type in {"none", "unanswerable"}:
        acc = contains_abstention_phrase(pred_n)
    elif answer_type in {"numeric", "percentage"}:
        p = first_number(pred_n)
        g = first_number(gold_n)
        acc = p is not None and g is not None and relative_error(p, g) <= 0.05
    elif answer_type == "single_choice":
        acc = first_option_label(pred) == first_option_label(gold)
    else:
        acc = (gold_n in pred_n) or norm_lev_sim(pred_n, gold_n) >= 0.8

    return {
        "acc": int(acc),
        "f1": token_f1(pred_n, gold_n),
        "anls": anls(pred, gold),
    }


def aggregate_scores(examples):
    scored = [evaluate_prediction(e.response, e.gold, e.answer_type) for e in examples]
    overall = mean_metrics(scored)
    by_reasoning = group_and_average(scored, key="reasoning_type")
    by_modality = group_and_average(scored, key="evidence_modality")
    by_scope = group_and_average(scored, key="document_scope")
    by_length = group_and_average(scored, key="context_or_evidence_bin")
    return overall, by_reasoning, by_modality, by_scope, by_length
\end{lstlisting}

\paragraph{Scoring details.}
\texttt{normalize} lowercases text, removes common answer prefixes, drops articles, strips most punctuation while retaining decimal points, hyphens, and percent signs, and collapses whitespace. Token F1 is set-based overlap after normalization. ANLS uses normalized Levenshtein similarity with a $0.5$ threshold and does not apply the full normalization step, so punctuation and surface form still affect the distance.

\section{Reasoning-Type and Cross-Document Case Studies}
\label{app:reasoning_case_studies}

\begin{figure}[p]
    \centering
    \includegraphics[width=\linewidth,height=0.94\textheight,keepaspectratio]{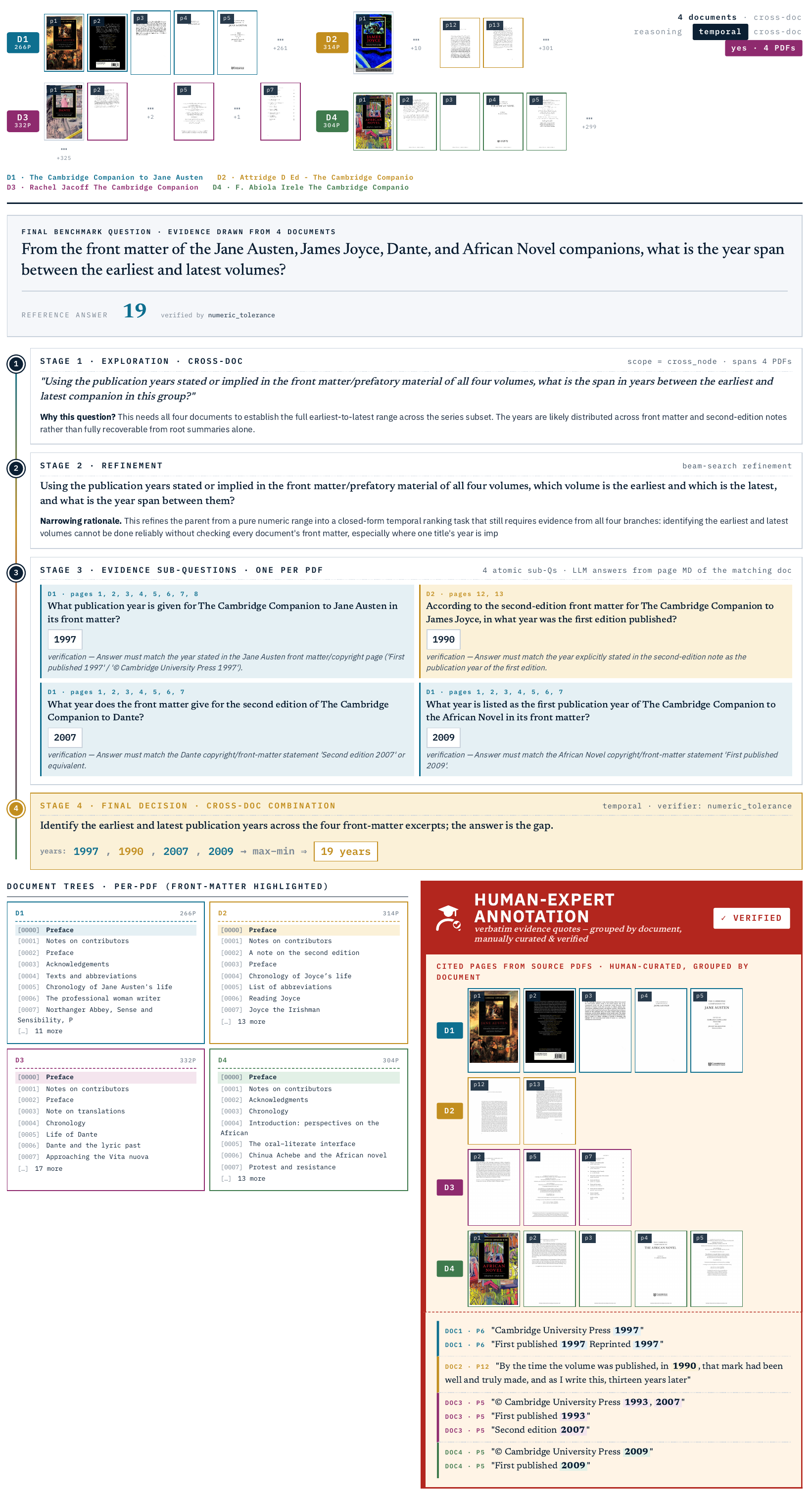}
    \caption{Case study: Cross-document temporal.}
    \label{fig:case_cross_doc_temporal}
\end{figure}

\begin{figure}[p]
    \centering
    \includegraphics[width=\linewidth,height=0.94\textheight,keepaspectratio]{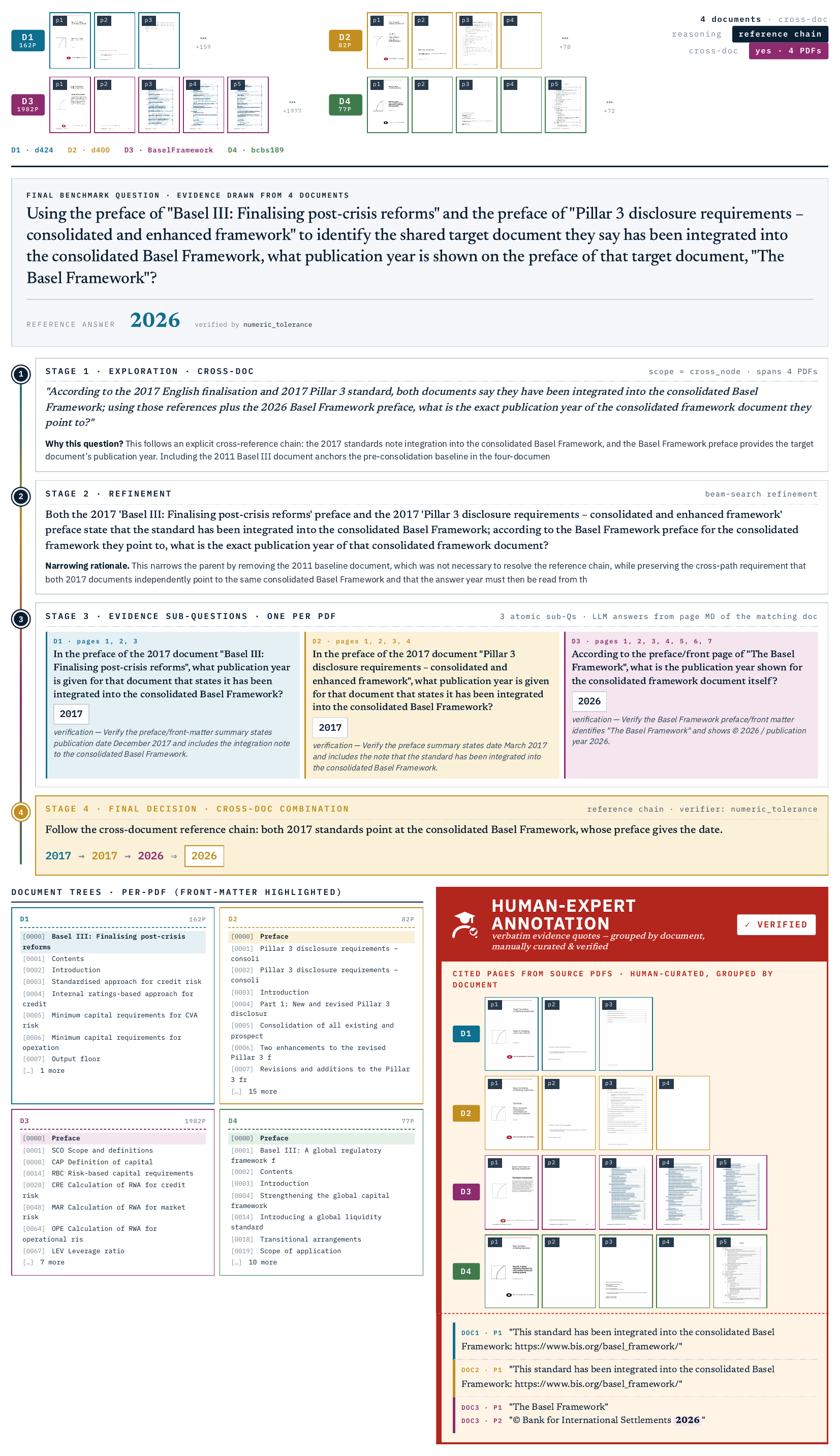}
    \caption{Case study: Cross-document reference chain.}
    \label{fig:case_cross_doc_reference_chain}
\end{figure}

\begin{figure}[p]
    \centering
    \includegraphics[width=\linewidth,height=0.94\textheight,keepaspectratio]{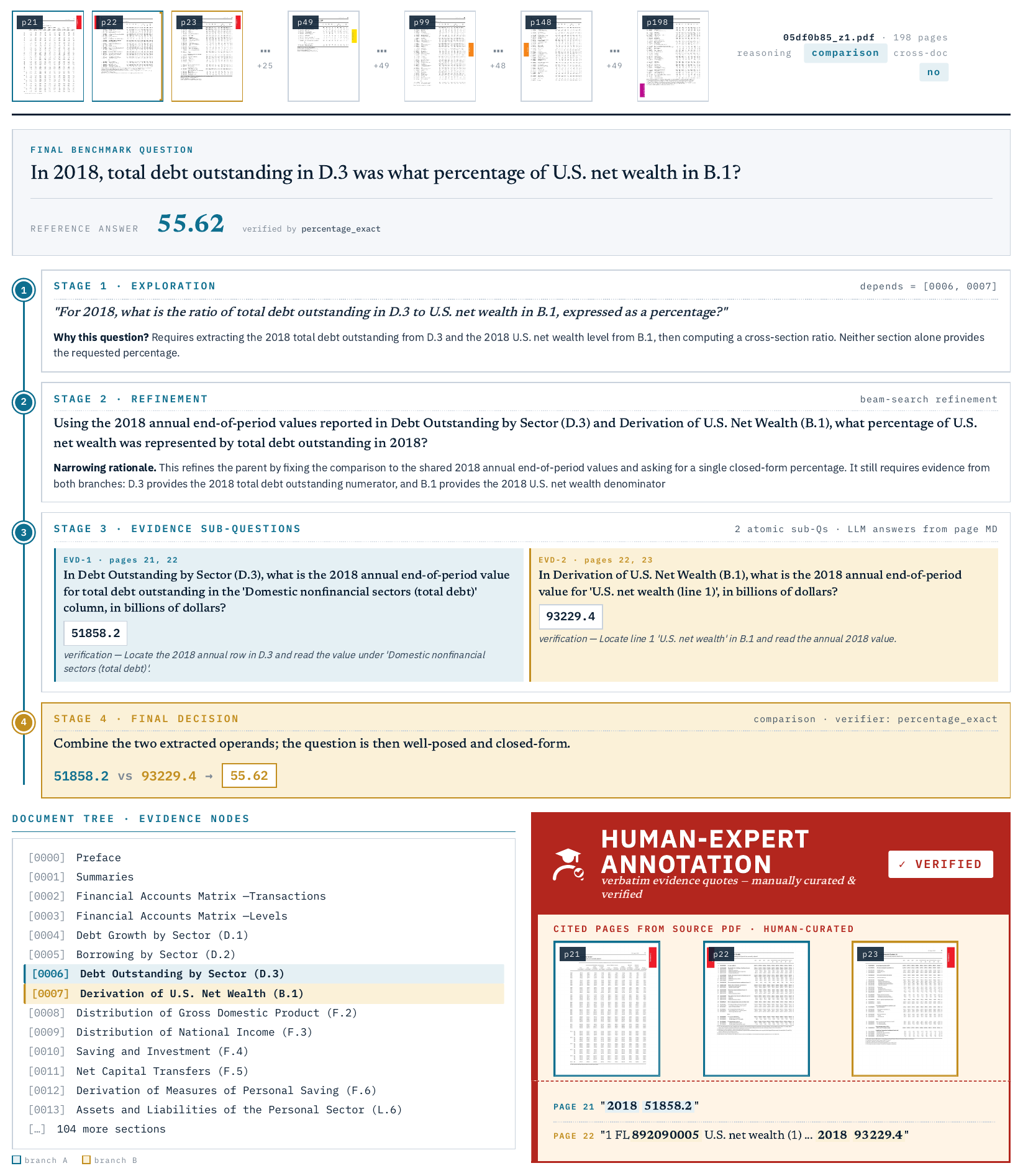}
    \caption{Case study: Comparison.}
    \label{fig:case_comparison}
\end{figure}

\begin{figure}[p]
    \centering
    \includegraphics[width=\linewidth,height=0.94\textheight,keepaspectratio]{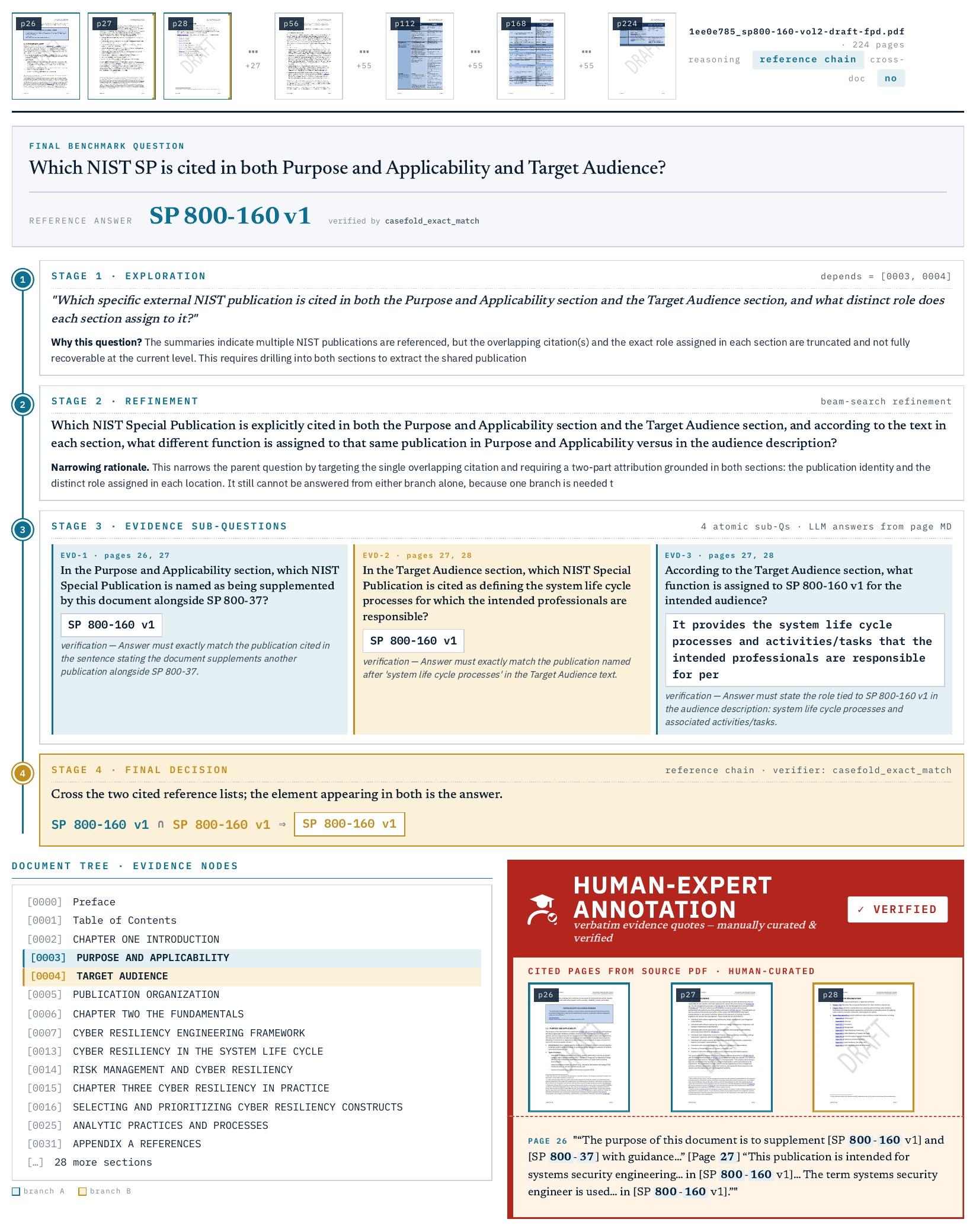}
    \caption{Case study: Reference chain.}
    \label{fig:case_reference_chain}
\end{figure}

\begin{figure}[p]
    \centering
    \includegraphics[width=\linewidth,height=0.94\textheight,keepaspectratio]{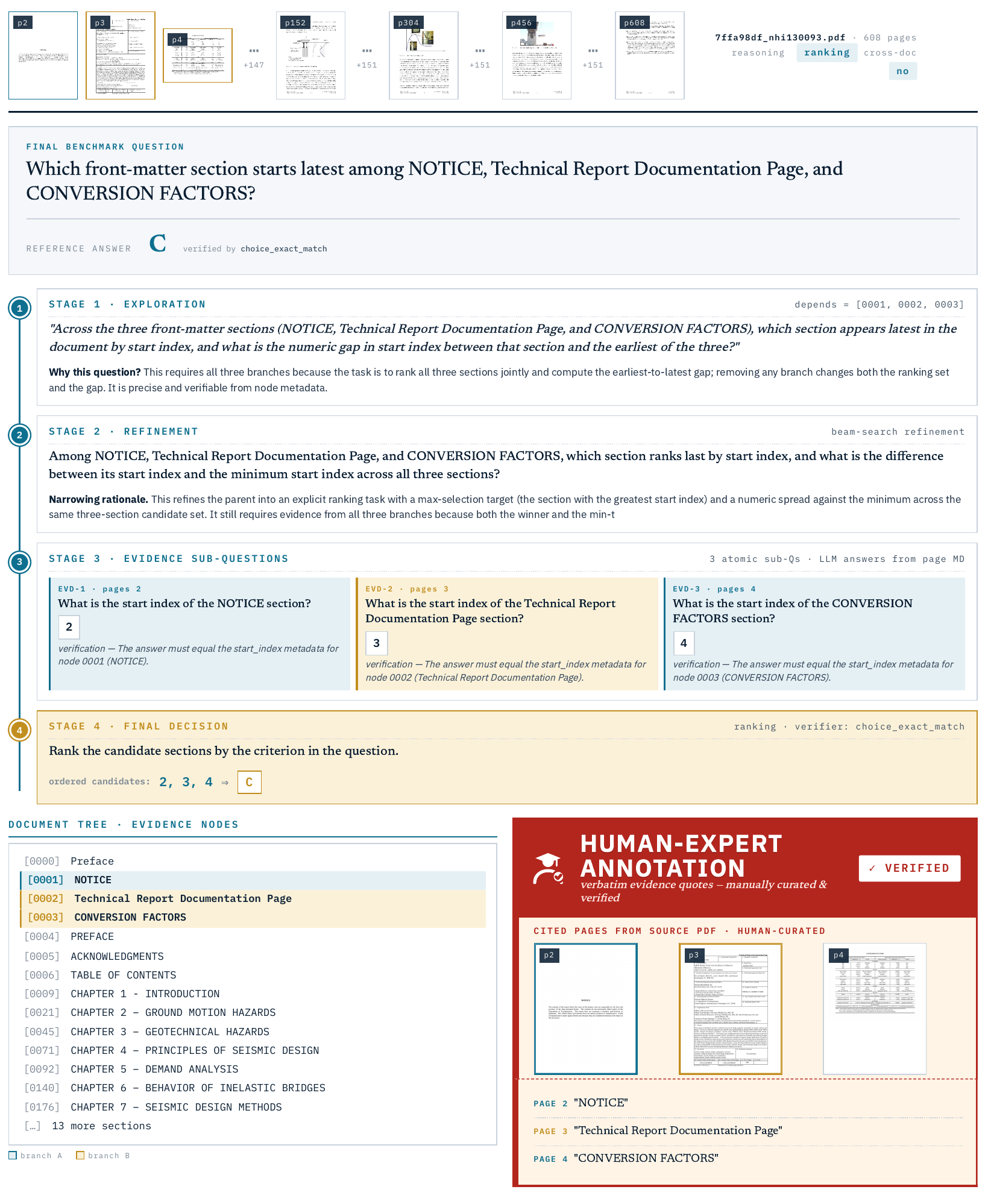}
    \caption{Case study: Ranking.}
    \label{fig:case_ranking}
\end{figure}

\begin{figure}[p]
    \centering
    \includegraphics[width=\linewidth,height=0.94\textheight,keepaspectratio]{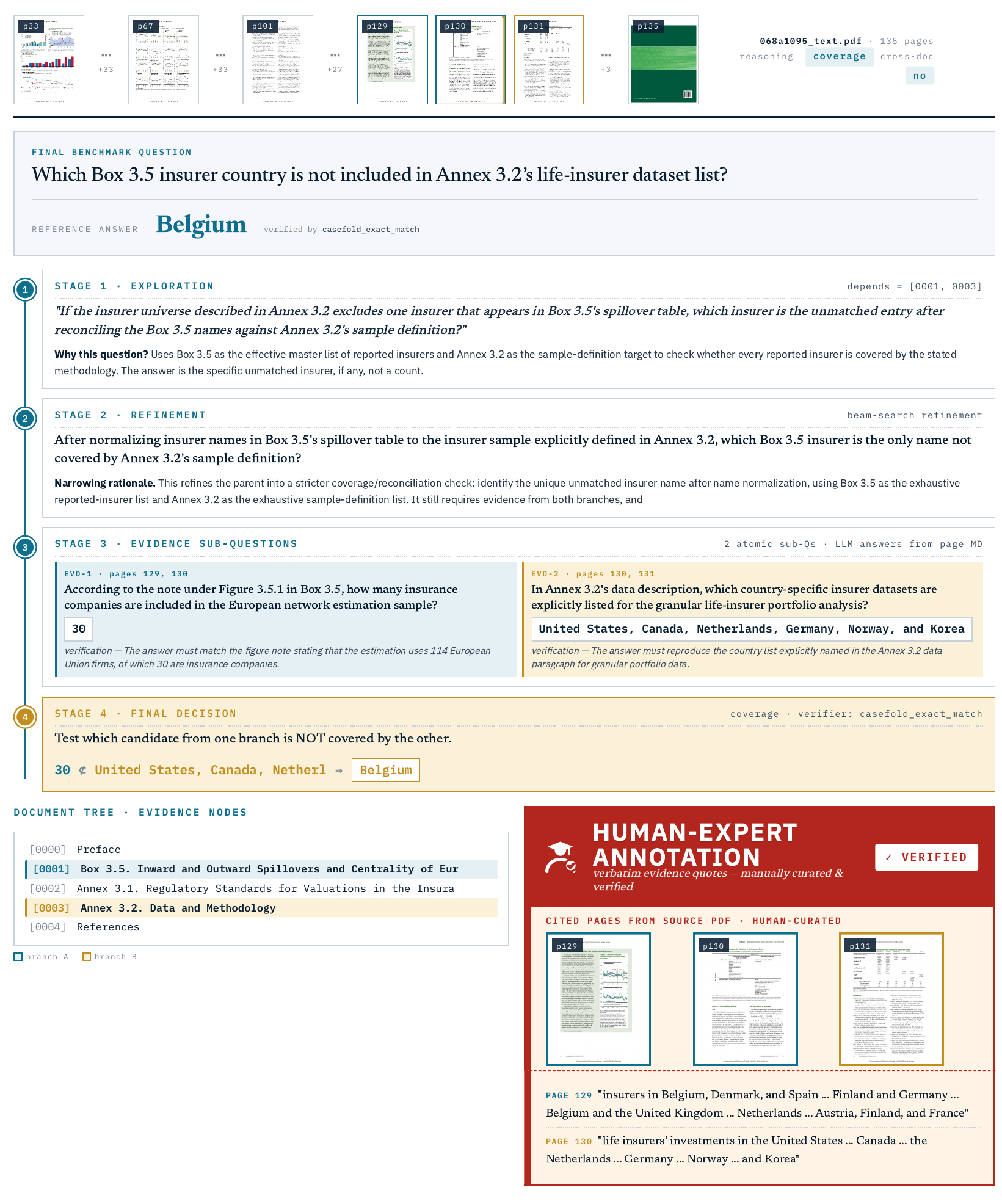}
    \caption{Case study: Coverage.}
    \label{fig:case_coverage}
\end{figure}

\begin{figure}[p]
    \centering
    \includegraphics[width=\linewidth,height=0.94\textheight,keepaspectratio]{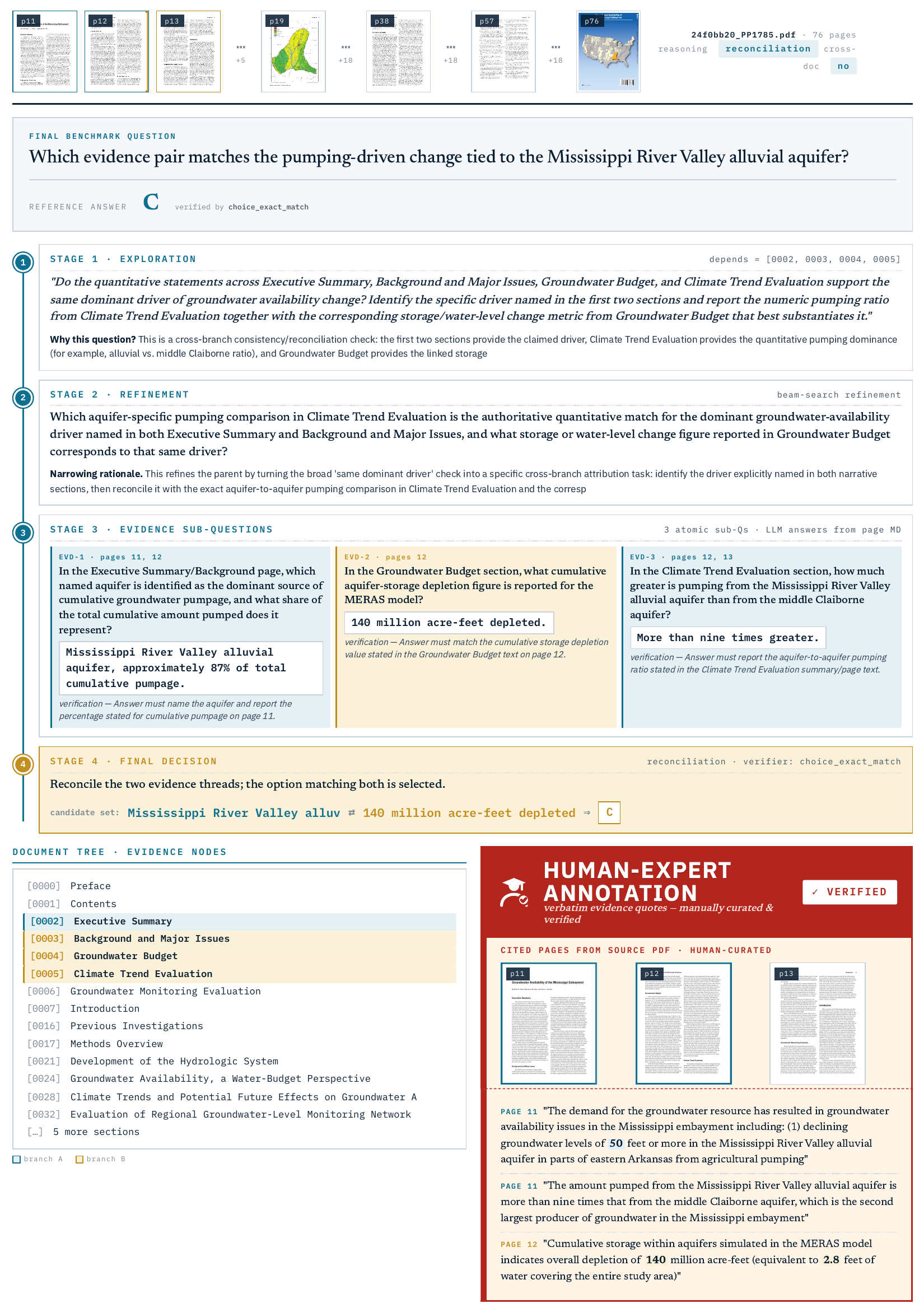}
    \caption{Case study: Reconciliation.}
    \label{fig:case_reconciliation}
\end{figure}

\begin{figure}[p]
    \centering
    \includegraphics[width=\linewidth,height=0.94\textheight,keepaspectratio]{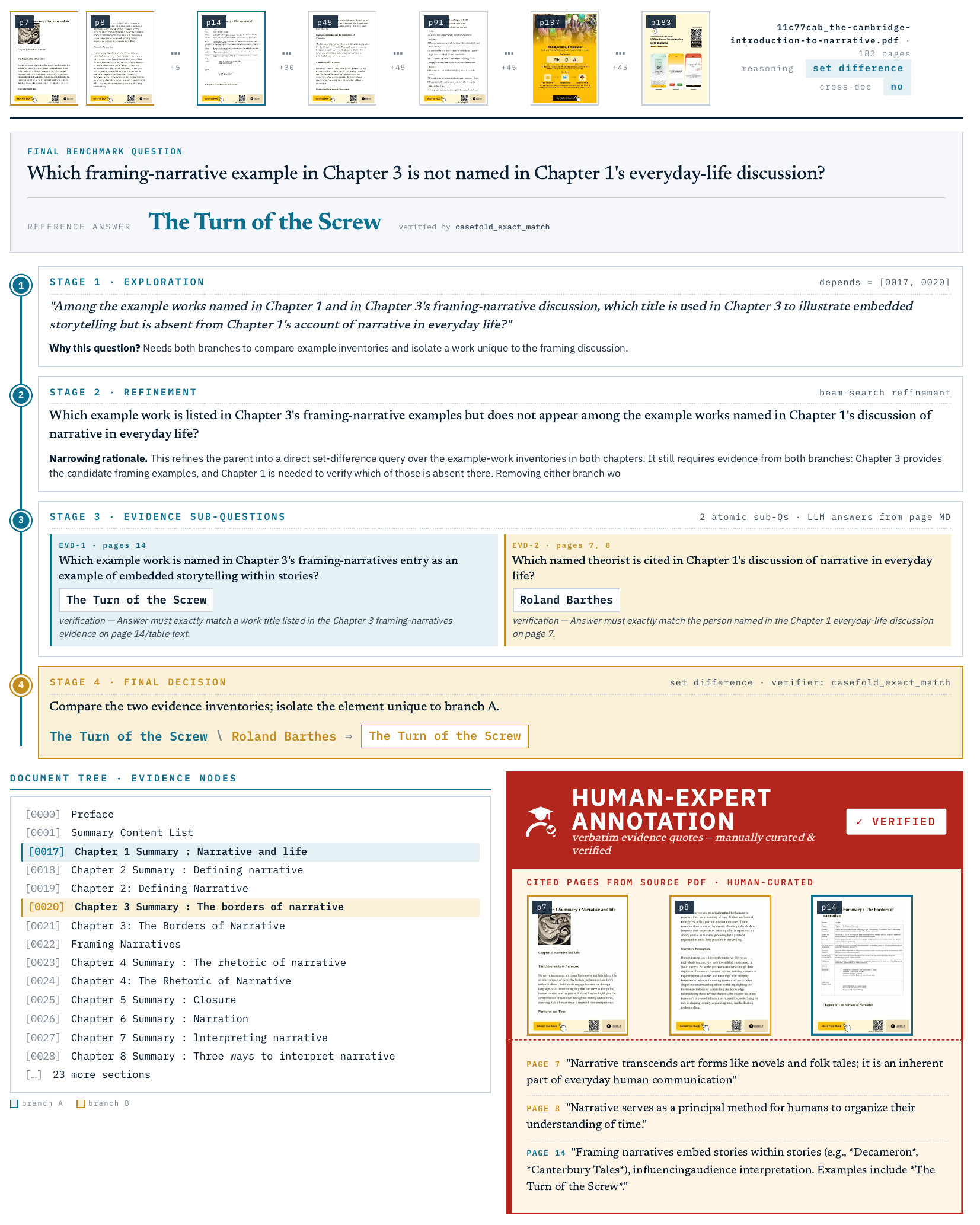}
    \caption{Case study: Set difference.}
    \label{fig:case_set_difference}
\end{figure}

\begin{figure}[p]
    \centering
    \includegraphics[width=\linewidth,height=0.94\textheight,keepaspectratio]{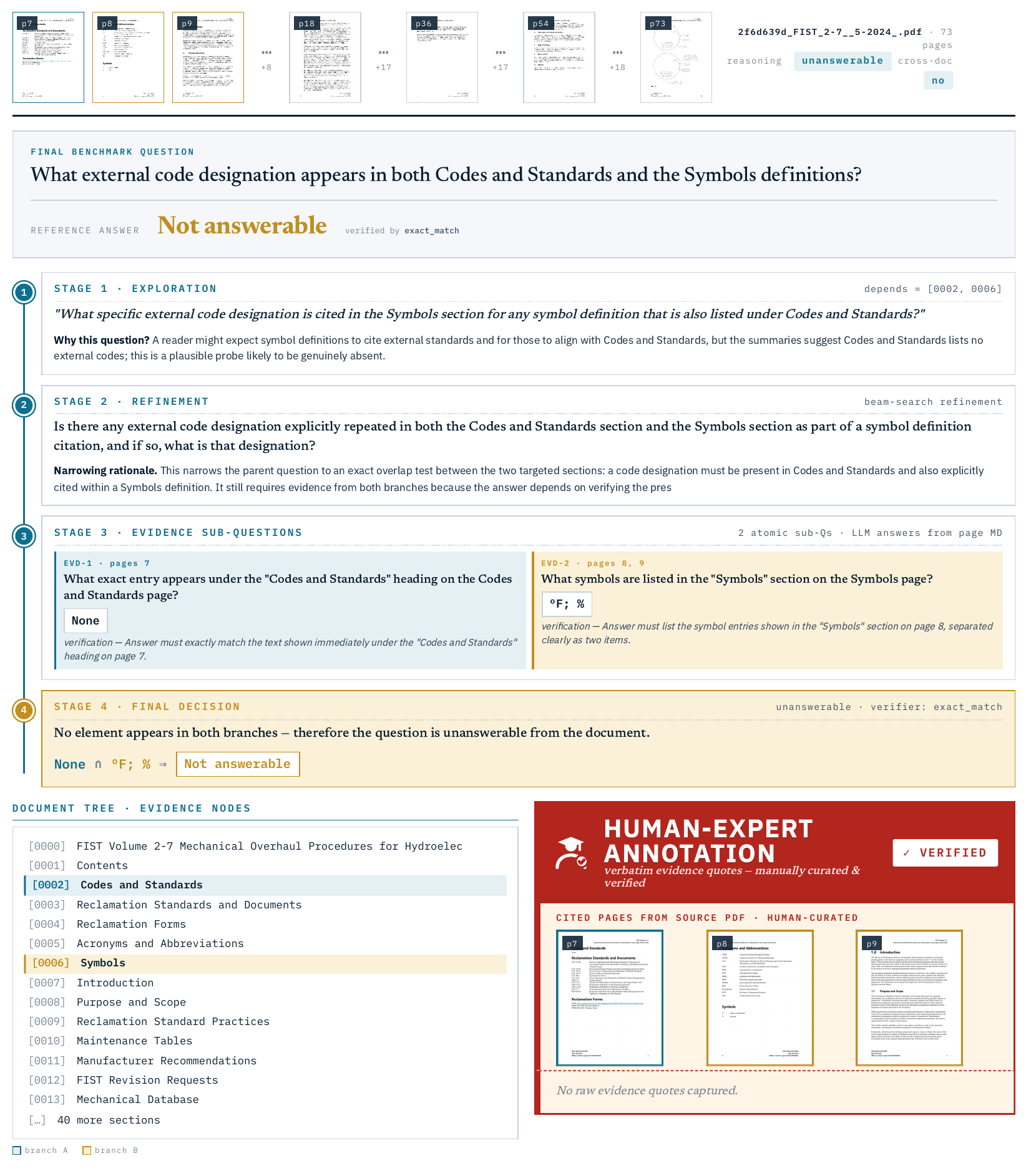}
    \caption{Case study: Unanswerable.}
    \label{fig:case_unanswerable}
\end{figure}

\begin{figure}[p]
    \centering
    \includegraphics[width=\linewidth,height=0.94\textheight,keepaspectratio]{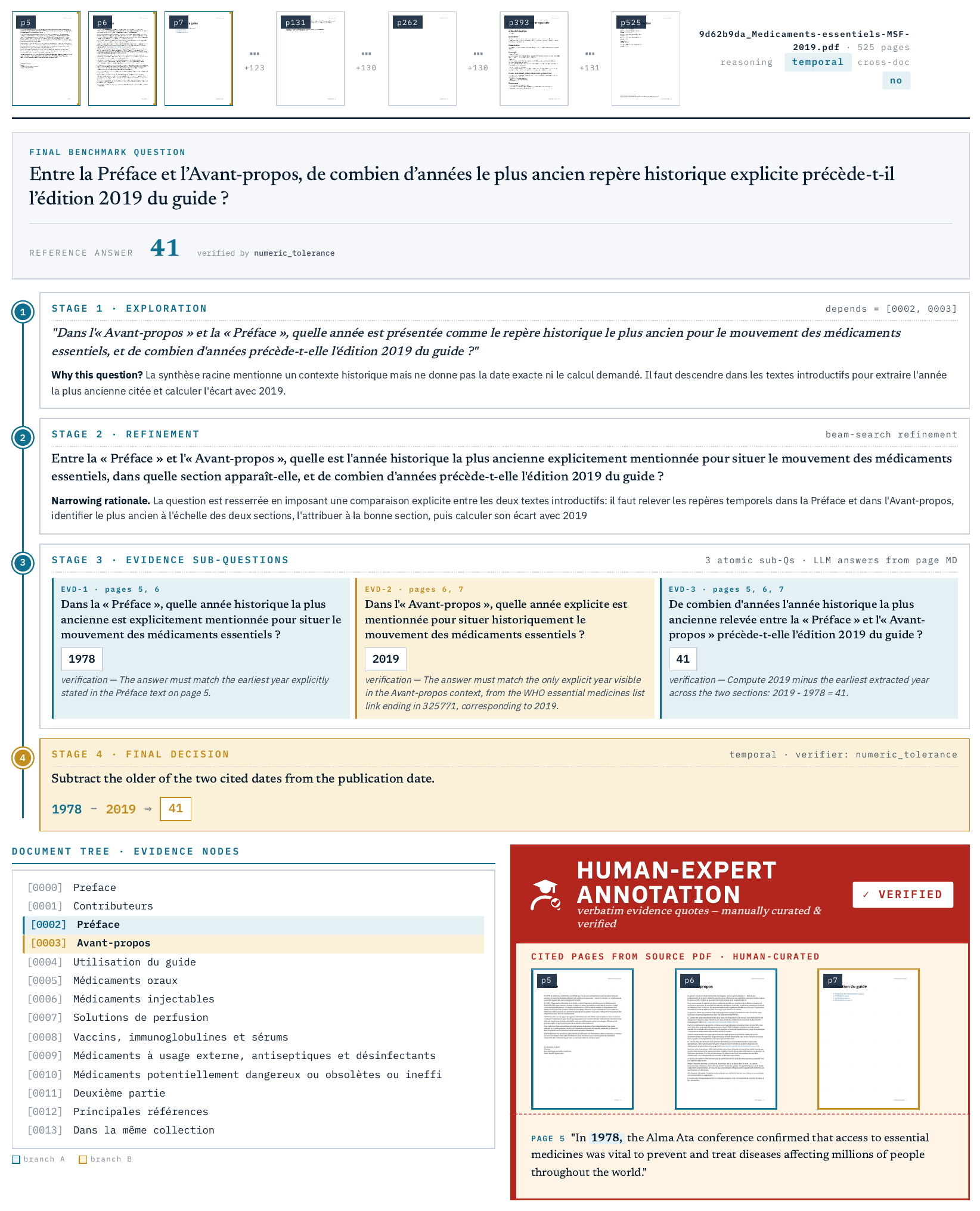}
    \caption{Case study: Temporal.}
    \label{fig:case_temporal}
\end{figure}

\begin{figure}[p]
    \centering
    \includegraphics[width=\linewidth,height=0.94\textheight,keepaspectratio]{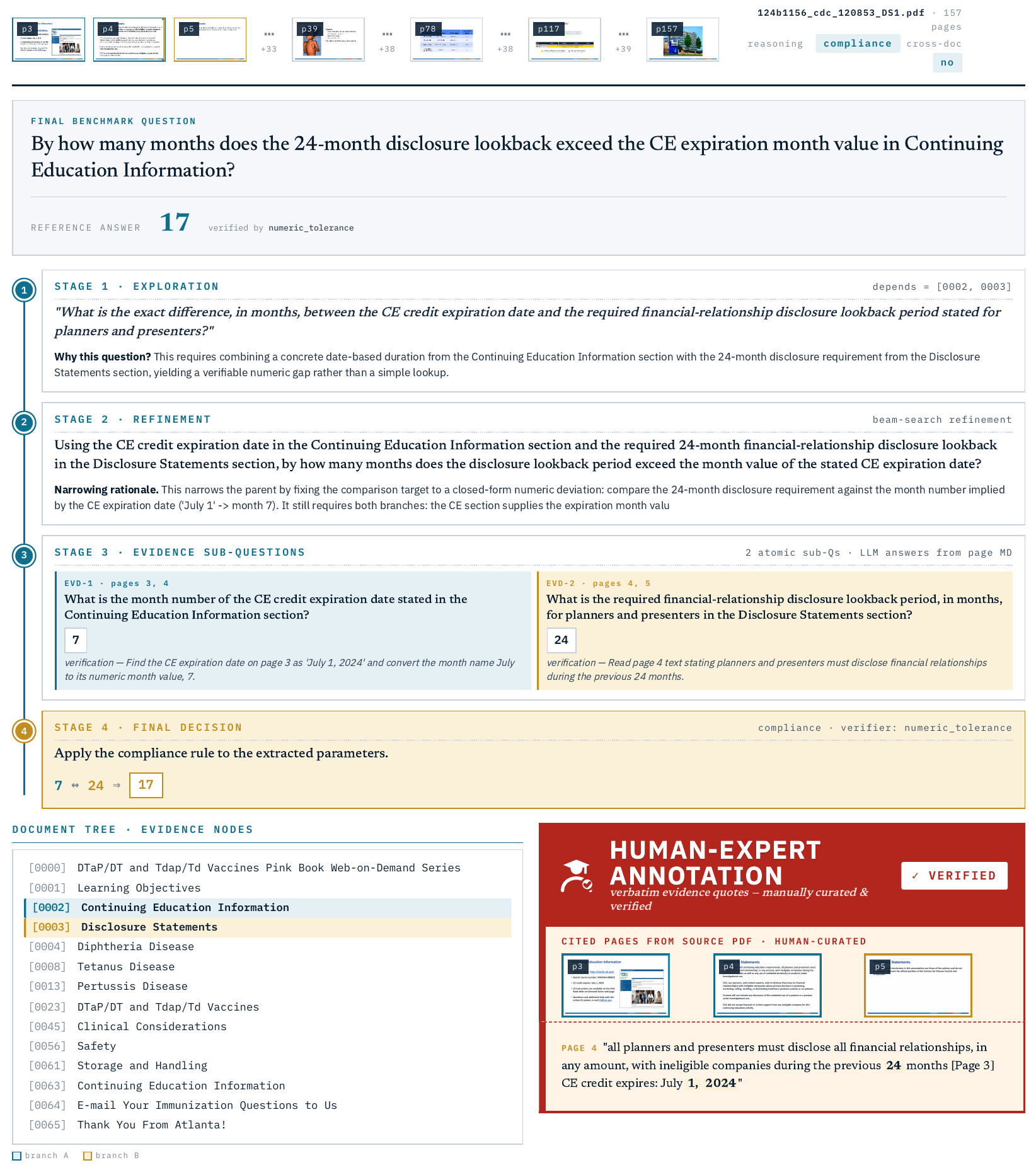}
    \caption{Case study: Compliance.}
    \label{fig:case_compliance}
\end{figure}

\begin{figure}[p]
    \centering
    \includegraphics[width=\linewidth,height=0.94\textheight,keepaspectratio]{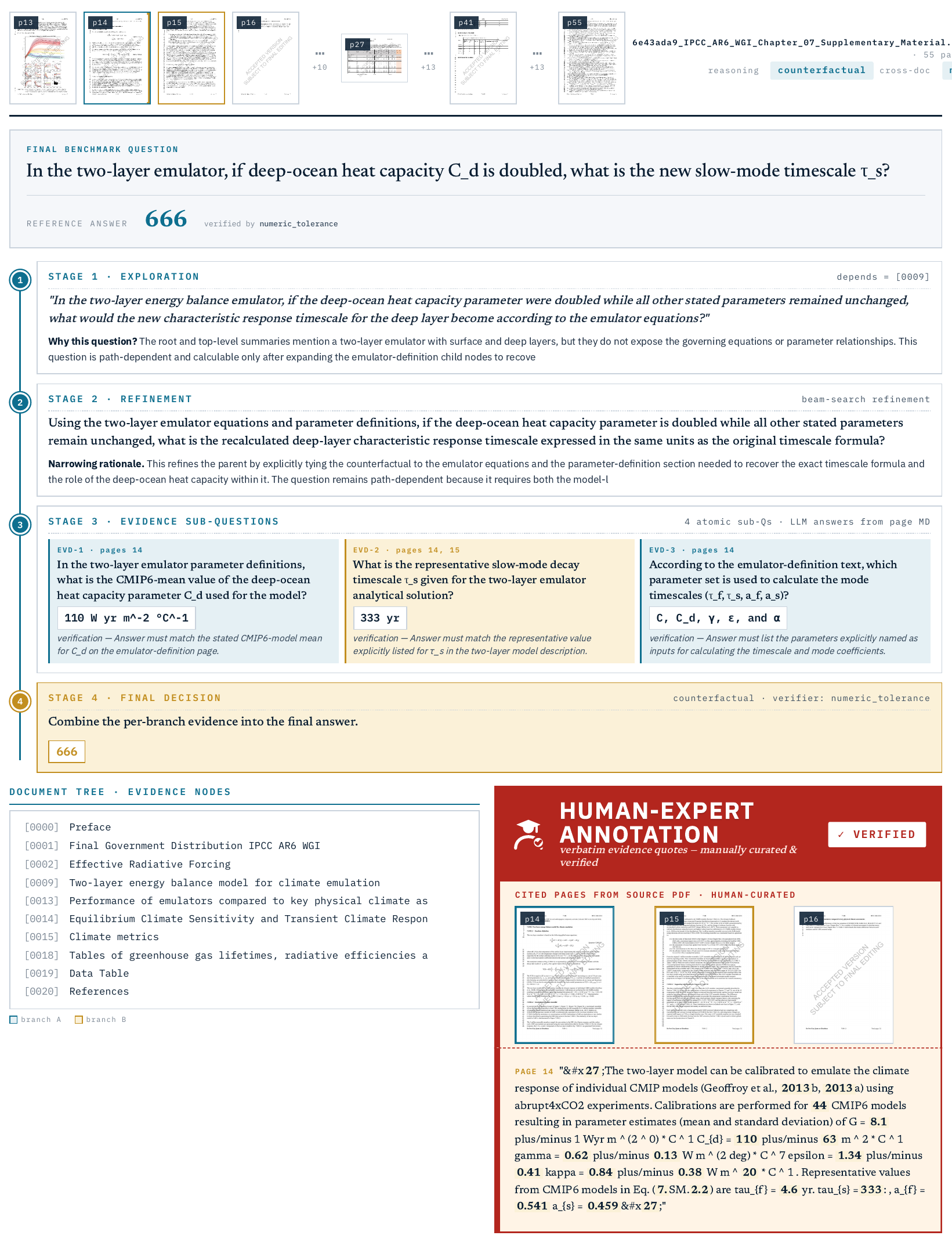}
    \caption{Case study: Counterfactual.}
    \label{fig:case_counterfactual}
\end{figure}

\begin{figure}[p]
    \centering
    \includegraphics[width=\linewidth,height=0.94\textheight,keepaspectratio]{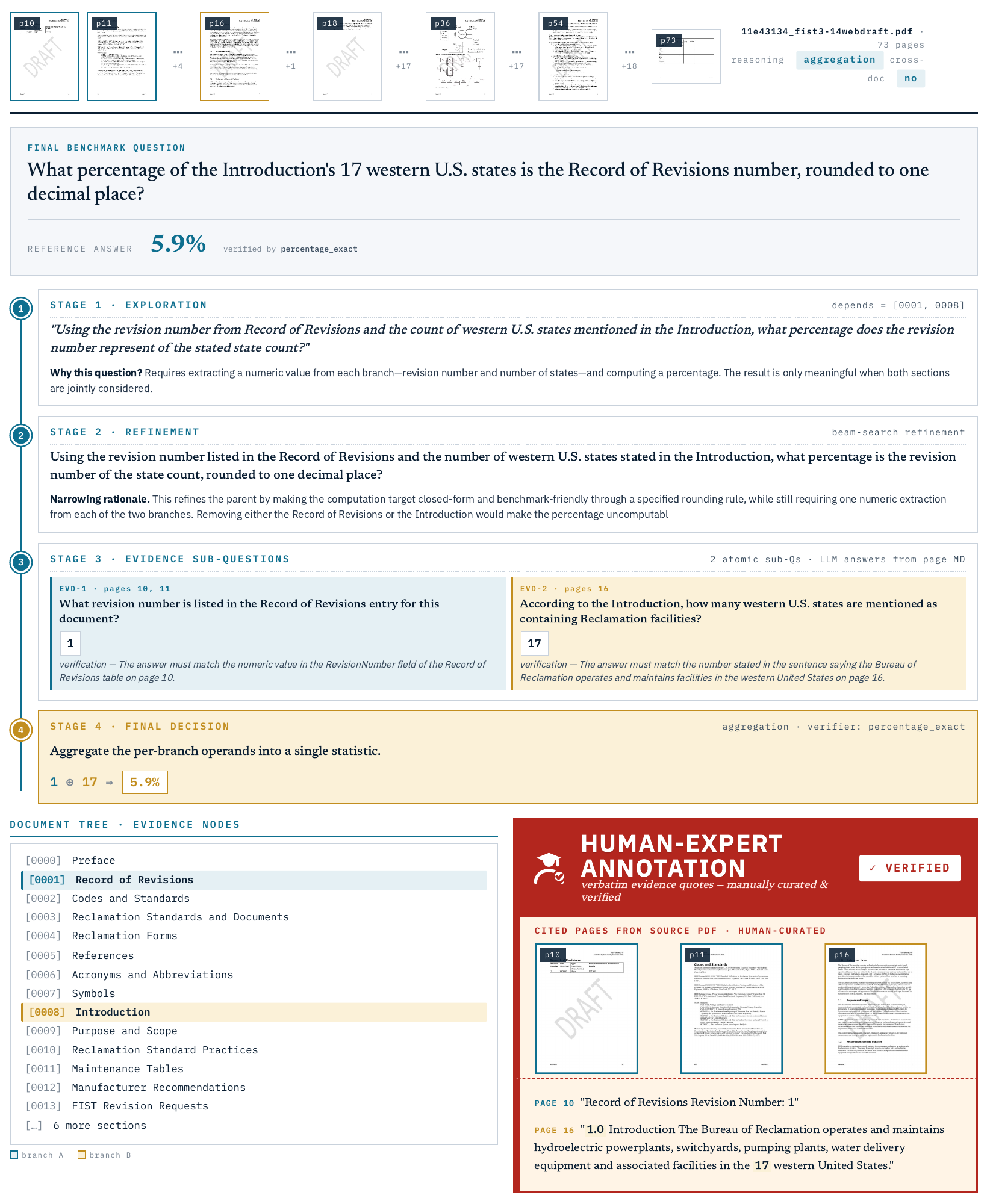}
    \caption{Case study: Aggregation.}
    \label{fig:case_aggregation}
\end{figure}

\begin{figure}[p]
    \centering
    \includegraphics[width=\linewidth,height=0.94\textheight,keepaspectratio]{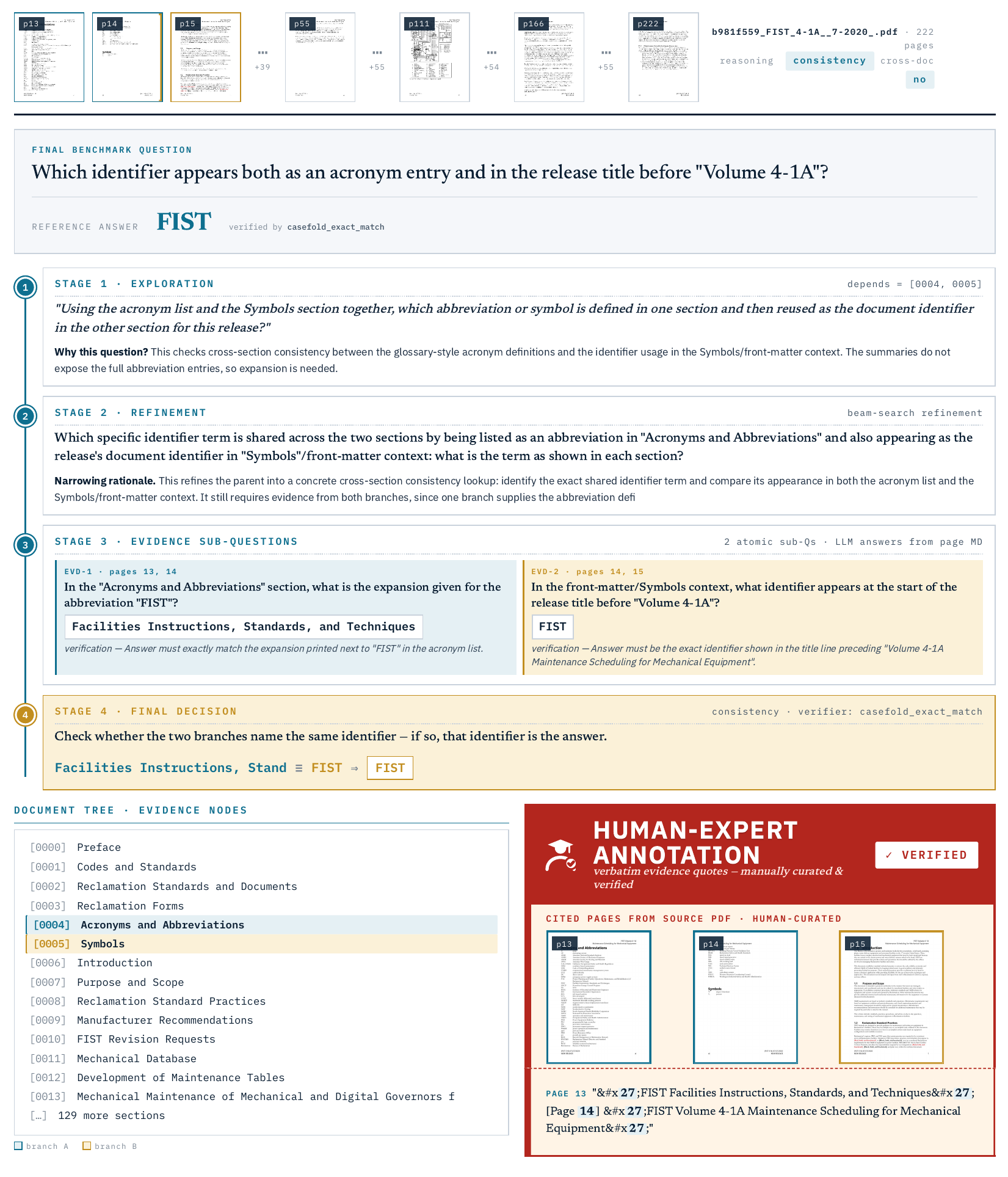}
    \caption{Case study: Consistency.}
    \label{fig:case_consistency}
\end{figure}

Figures~\ref{fig:case_cross_doc_temporal}--\ref{fig:case_consistency} show cases from the final human-verified benchmark. Cross-document cases appear first, followed by reasoning-type cases. Each case comes from the same construction pipeline used for the benchmark. Documents are parsed into page text, layout, tables, figures, and page images. Candidate questions are generated from document branches and assigned a reasoning type. Automatic checks filter candidates for answer format, leave-one-branch sensitivity, and evidence consistency. The remaining examples are sent to human verification. The figures show final verified examples rather than raw generated candidates. The question, typed answer, reasoning label, evidence pages, and marked evidence spans are the result of human review. Human annotators inspected the source pages, corrected or removed unsupported evidence, and verified that the answer follows from the marked support. They also resolved ambiguous cases and confirmed whether cross-document cases require evidence from multiple PDFs. These examples show how human evidence verification turns automatically proposed questions into reliable benchmark examples.

\newpage

\end{document}